\documentclass[10pt,journal,compsoc]{IEEEtran}

\ifCLASSOPTIONcompsoc
  \usepackage[nocompress]{cite}
\else
  \usepackage{cite}
\fi

\ifCLASSINFOpdf
\else
\fi

\usepackage{colortbl}

\usepackage{times}
\usepackage{epsfig}
\usepackage{graphicx}
\usepackage{amsmath}
\usepackage{amssymb}
\usepackage{dblfloatfix}
\usepackage{array,multirow,graphicx}
\usepackage{multirow}
\usepackage{diagbox}
\usepackage{gensymb}
\usepackage{booktabs}

\usepackage{xcolor}

\definecolor{marron}{rgb}{0.5,0.3,0}
\definecolor{vertFonce}{rgb}{0,0.69,0.07}

\newcommand{\romain}[1]{\textcolor{black}{#1}}
\newcommand{\benjamin}[1]{\textcolor{black}{#1}} 
\newcommand{\marius}[1]{\textcolor{black}{#1}}
\newcommand{\pierre}[1]{\textcolor{black}{#1}}

\usepackage{tikz}

\usepackage{pifont}

\newcommand{\cc}[1]{%
 \ifnum #1 = 0 %
     \cellcolor{red!0} %
  \else%
  \ifnum #1 < 2 %
     \cellcolor{red!5} %
  \else%
    \ifnum #1 < 4 %
        \cellcolor{red!10} %
    \else %
        \ifnum #1 < 6 %
            \cellcolor{red!20} %
        \else %
            \ifnum #1 < 8 %
                \cellcolor{red!30} %
            \else %
                \ifnum #1 < 10 %
                    \color{white}%
                    \cellcolor{red!40} %
                \else %
                    \ifnum #1 > 9 %
                        \color{white}%
                        \cellcolor{red!50} %
                        \fi%
                    \fi%
                \fi%
            \fi%
        \fi%
    \fi%
    \fi%
 }
 
 \newcommand{\ccc}[1]{%
 \ifnum #1 = 0 %
     \cellcolor{green!0} %
  \else%
  \ifnum #1 < 2 %
     \cellcolor{green!5} %
  \else%
    \ifnum #1 < 4 %
        \cellcolor{green!10} %
    \else %
        \ifnum #1 < 6 %
            \cellcolor{green!20} %
        \else %
            \ifnum #1 < 8 %
                \cellcolor{green!30} %
            \else %
                \ifnum #1 < 10 %
                    \color{white}%
                    \cellcolor{green!40} %
                \else %
                    \ifnum #1 > 9 %
                        \color{white}%
                        \cellcolor{green!50} %
                        \fi%
                    \fi%
                \fi%
            \fi%
        \fi%
    \fi%
    \fi%
 }
 
\usepackage{enumitem}
\setlist[itemize]{align=parleft,left=0pt..1em}

\begin{document}

\title{Impact of Facial Landmark Localization on Facial Expression Recognition}

\author{Romain Belmonte, Benjamin Allaert, Pierre Tirilly, Ioan Marius Bilasco, Chaabane Djeraba, and Nicu Sebe,~\IEEEmembership{Senior Member,~IEEE}}

\markboth{IEEE Transactions on Affective Computing}%
{Shell \MakeLowercase{\textit{et al.}}: Bare Demo of IEEEtran.cls for Computer Society Journals}

\IEEEtitleabstractindextext{%
\begin{abstract}
Although facial landmark localization (FLL) approaches are becoming increasingly accurate for characterizing facial regions, one question remains unanswered: what is the impact of these approaches on subsequent related tasks? In this paper, the focus is put on facial expression recognition (FER), where facial landmarks are used for face registration, which is a common usage. Since the most used datasets for facial landmark localization do not allow for a proper measurement of performance according to the different difficulties (e.g., pose, expression, illumination, occlusion, motion blur), we also quantify the performance of recent approaches in the presence of head pose variations and facial expressions. Finally, a study of the impact of these approaches on FER is conducted. We show that the landmark accuracy achieved so far optimizing the conventional Euclidean distance does not necessarily guarantee a gain in performance for FER. To deal with this issue, we propose a new evaluation metric for FLL adapted to FER. 
\end{abstract}

}

\maketitle

\IEEEdisplaynontitleabstractindextext

\IEEEpeerreviewmaketitle

\IEEEraisesectionheading{\section{Introduction}\label{sec:introduction}}

\IEEEPARstart{D}{espite} continuous progress in facial expression recognition (FER), many studies still focus on near-frontal faces \cite{li2020deep}. As a result, most approaches perform poorly when head pose variations occur (e.g., in video surveillance data). Recent work considers the whole range of head poses under uncontrolled conditions and this issue receives more and more attention \cite{zhang2018joint}. 

Facial landmark localization (FLL) approaches have proven their effectiveness in identifying the facial components of the face (eyes, nose, and mouth) \cite{deng2018menpo}. Once located, these landmarks can be used to deal with head pose variations \cite{hassner2015effective} and support FER as well as various facial analysis tasks \cite{allaert2018advanced}. For example, the face can be registered in order to guarantee stable locations for the major facial components across images and minimize the variations in scale, rotation, and position. It makes it then easier to extract discriminative features to characterize the face.

\begin{figure}[!h]
\centering
\includegraphics[width=\columnwidth]{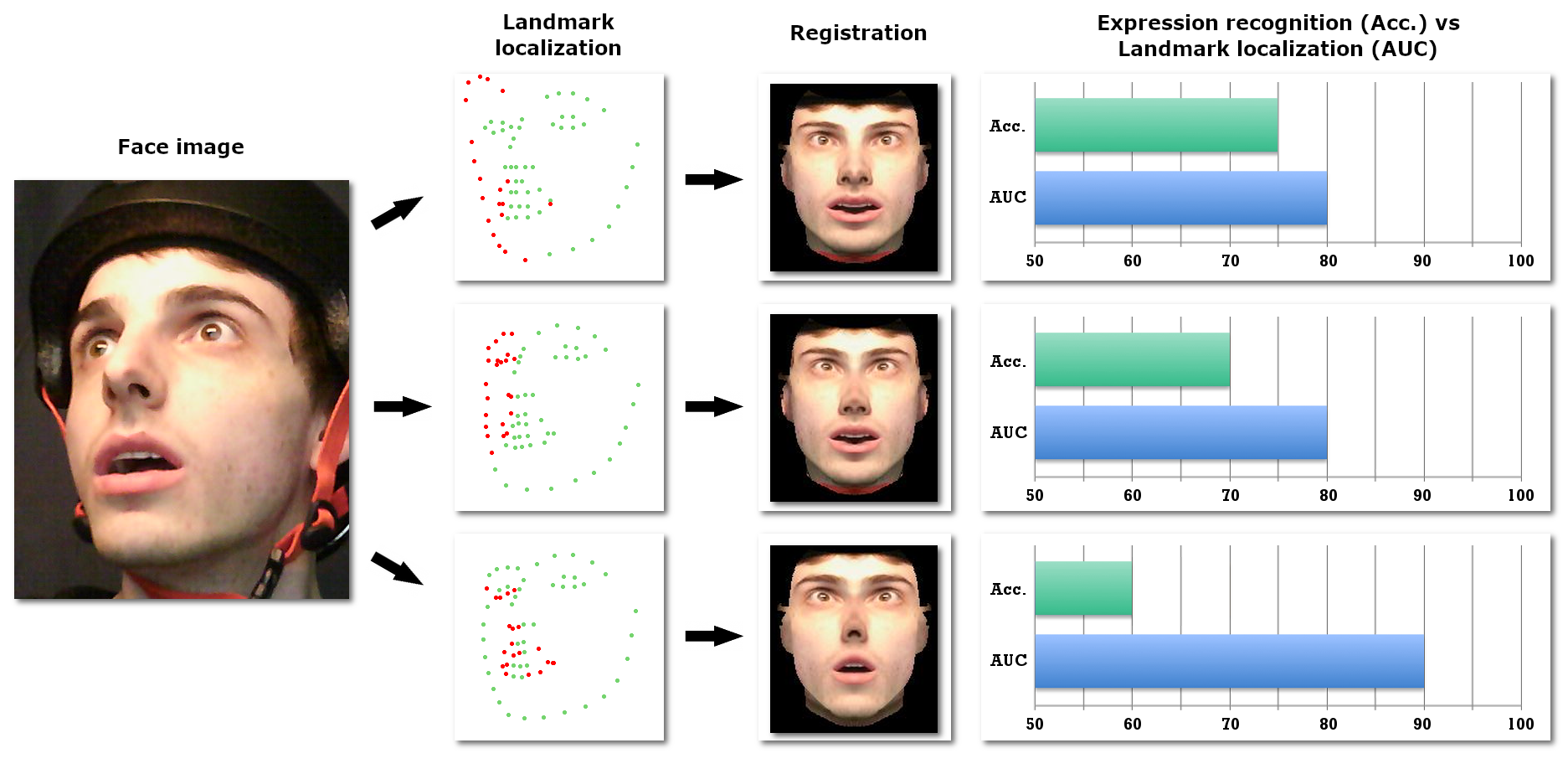} 
\caption{Comparison of the ability of different FLL approaches to provide meaningful facial geometry information when used to register a face. On the right-hand side of the image, we illustrate the registered face using the landmarks provided by some of the approaches considered in this paper. The graphs present a comparison of FLL performance (measured as the Area Under the Curve - AUC) and FER performance (measured as the accuracy - Acc.). A higher average AUC does not guarantee that the corresponding localization is better suited to recognize facial expressions.} 
\label{fig:landmarks}
\end{figure}

Over the years, more robust approaches for FLL have been proposed, which address a wide range of difficulties (head poses, facial expressions, illumination, etc.). However, due to the limits of popular datasets (e.g., 300W \cite{sagonas2016300}) in terms of contextual annotations, we lack a good understanding of the performances of current approaches. Evaluations are generally limited to overall performance. The performance according to the different difficulties cannot be properly quantified. Besides, evaluation metrics are mainly based on the Euclidean distance, which is not really relevant to FER or other subsequent tasks. There is no guarantee that an approach providing more accurate detections in terms of Euclidean distance always leads to better performance when used for FER, as illustrated in Figure \ref{fig:landmarks}. We show in this work that some landmarks have a greater impact on face registration. More significant geometric deformations can be induced if the localization fails on important landmarks such as the rigid ones.

The main contributions of this paper are as follows.
\begin{itemize}
    \item We quantify simultaneously the performance of current FLL approaches in presence and in absence of head pose variations and facial expressions. This allows us to get a clear view of their strengths and weaknesses, so that we can interpret the results of the following experiments more accurately.
    \item We study the impact of FLL on FER. Among other things, this study highlights the respective importance of landmarks for FER. Based on our observations, we provide some suggestions on how to improve FLL considering FER.
    \item \marius{We propose two  changes to the FLL evaluation metric, to improve the correlation between the FLL precision and the FER accuracy when FLL approaches are used to register faces.}
\end{itemize}  

The paper is structured as follows. Section \ref{bscope} presents the evolution of recent approaches for FLL and FER and highlights the main objectives of our work. Section \ref{dataset} introduces the dataset and FLL approaches used in the experiments. Section \ref{landmark} provides an evaluation of the performances of the selected FLL approaches in the presence of head pose variations and facial expressions. Section \ref{expression} investigates the ability of these approaches to capture meaningful facial geometry information when used for FER. \marius{{Section \ref{metrics} investigates new FLL evaluation metrics coping with FER analysis challenges and improving the correlation between FLL precision and FER accuracy.}} Finally, we discuss the results and future work in Section \ref{conclusion}.



\section{Background and scope}
\label{bscope}

\marius{This section presents challenges related to FER and recent FLL approaches. We present an overview of the literature on FER that shows that under different  capturing  conditions  (i.e.,  in the  presence  or  absence  of  head  movements,  head  poses,  facial expressions), FLL plays an important role in the process. Then, we report recent advances in the field of FLL. Finally, the scope of the paper is defined, including our positioning, the issues raised by this work and how we address them.}




\subsection{Facial expression recognition}
Since the majority of work is still done under controlled conditions, we first present this line of work and then review the work intended for uncontrolled conditions. For the latter, we distinguish between landmark-based and landmark-free approaches. 

\subsubsection{Under controlled conditions}
Skin deformations induced by facial muscles characterize facial expressions. Different approaches exist to encode facial deformations: appearance features, geometry features, or both. Many appearance features have been proposed such as Local Binary Patterns (LBP) \cite{lbp}. They provide decent results for macro facial deformations, especially when they are computed on apex frames (i.e., the frames of a video that depict the expressions at their highest intensity). By relying on spatial features only, the dynamics of facial expressions is not leveraged, which can limit the performances at non-apex frames or in the presence of subtle expressions. 

Psychological experiments \cite{bassili1979emotion} showed that facial expressions are identified more accurately in image sequences. A dynamic extension of LBP, called Local Binary Pattern on Three Orthogonal Plans (LBP-TOP), has been proposed \cite{lbptop}. However, optical flow is one of the most widely used solutions \cite{allaert2018advanced}. Although temporal approaches tend to provide better performances than static approaches, they are very sensitive to the noise caused by facial deformation artifacts and head movements.

Various deep learning approaches have also been proposed. Pre-training and fine-tuning are commonly used to deal with small datasets. Some work focuses on improving the ability of models to represent expressions when using such techniques \cite{ding2017facenet2expnet}. Handcrafted features, in conjunction with the raw image, can be used as input to add information and increase invariance, e.g., to scale, rotation, or illumination. Layers, blocks, and loss functions have been specifically designed for FER in order to encourage the learning of expression-related features \cite{hu2017learning}. For example, second-order statistics such as covariance have been used to capture regional distortions of facial landmarks \cite{acharya2018covariance}. Other well-known strategies have also been applied to FER, i.e., network ensembles, multitask networks, cascaded networks, and generative adversarial networks \cite{li2020deep, wang2020learning}. In order to extend these solutions to the temporal domain, aggregation strategies have been proposed, which aim to combine the outputs of static networks for each frame of a sequence \cite{kahou2013combining}. Samples of the same expression with different intensities can be used jointly as input during training to help address subtle expressions \cite{zhao2016peak}. Deep spatio-temporal networks based on recurrent neural networks \cite{le2015simple} and 3D convolution \cite{tran2015learning} are also used to encode expressions by exploiting both appearance information and motion information from images or landmarks \cite{jung2015joint}.

\subsubsection{Under uncontrolled conditions}
All these approaches have proven their effectiveness in characterizing facial expressions on static frontal faces, but they are rarely evaluated under uncontrolled conditions. In situations of natural interactions, facial expression analysis remains a complex problem. It requires the solutions to be invariant to head pose variations (i.e., in-plane and out-of-plane rotations) and large head displacements (i.e., large in-plane translations). To deal with this issue, two major approaches exist: landmark-based approaches and landmark-free approaches.

\vspace{3pt}

\noindent \pierre{\textbf{Landmark-based approches}} Landmark-based approaches \cite{zhang2018joint, hu2017learning} \cite{acharya2018covariance} are probably the most popular. 
They rely on facial landmarks to bring the face into a canonical configuration, typically a frontal head pose. Eye registration is the most popular strategy for near-frontal view. However, the eyes must be detected accurately in the first place. Extensions considering more landmarks are supposed to provide better stability when individual landmarks are poorly detected. Registration approaches based on 2D features \cite{schaefer2006image} are suitable for near-frontal facial expression analysis in the presence of limited head movements, but they encounter issues during occlusions and out-of-plane rotations. Registration based on 3D models can be used to generate natural facial images in a frontal head pose \cite{hassner2015effective}. Compared to 2D approaches, 3D approaches reduce facial deformation artifacts when facial expressions occur \cite{allaert2018impact}. \pierre{However, 3D approaches are not yet widely used. The lack of suitable data and available models made it difficult to include them in this study.} Since all these approaches are based on facial landmarks, an accurate localization of these landmarks is critical. Poor localization is likely to impact the entire recognition process.

\vspace{3pt}

\noindent \pierre{\textbf{Landmark-free approaches}} Few approaches use directly (without any preprocessing) the face crop obtained from face detection \cite{levi2015emotion}. Recently, a new way of addressing the problem has been proposed. It consists in the regression of a 3D morphable model (3DMM) directly from the image \cite{chang17expnet} using deep neural networks. 
\pierre{The facial expression is encoded as a vector of facial deformation coefficients.} These coefficients can then be used to classify the expression. This is referred to as facial expression regression. 3DMM coefficients could have a strong discriminative power while being less opaque than deep features. Recent work includes the release of a new dataset of 6,000 human facial videos, called Face-Vid \cite{koujan2020real}, to help work in this direction. This approach, being more direct, limits in principle the side effects of FLL and normalization. Due to the 3D representation of the face, better robustness to the difficulties encountered in uncontrolled conditions, such as occlusions, is also claimed. \pierre{Despite the theoretical advantages, this approach seems yet not efficient enough to take over landmark-based approaches. In the Appendix \ref{appendixa} we propose a comparison between ExpNet \cite{chang17expnet} and a landmark-based solution CovPool \cite{acharya2018covariance}. Under the experimental protocol used, the results show that : (1) the CovPool performances are higher and (2) even ExpNet can benefit from the registration phase.}


\subsection{Facial landmark localization}
\label{section:review:alignment}
Human-defined landmark schemes allow the components of the face to be explicitly characterized. In this work, we exploit facial landmarks for normalization purposes only, which is the most common usage. Note that normalization is not the only use of facial landmarks. For example, they can be used to produce geometric features \cite{jung2015joint}, to analyze motion from landmark trajectories \cite{zhang2017facial}, or to learn more specific features through attention mechanisms \cite{gan2020multiple}. Different landmark schemes with either few landmarks (e.g., only the rigid ones for each facial component) or with over a hundred landmarks (e.g., including the contours of the face and of each component) are used depending of the application (e.g., human-computer interaction, motion capture, FER) \cite{sagonas2016300}. The 68-landmark scheme is usually considered to be suitable for many applications with a reasonable trade-off between annotation time and informative content. 

\subsubsection{Localization in still images}

The majority of FLL approaches are based on cascaded regression \cite{Zafeiriou_2017_CVPR_Workshops}. It is a coarse-to-fine strategy that consists in progressively updating the positions of landmarks through regression functions learned directly from features representing the appearance of the face. Today, feature extraction and regression are trained jointly using deep neural networks. Two main architectures can be distinguished: a) networks that directly regress landmark coordinates using a final fully connected layer, and b) fully convolutional networks (i.e., without any fully connected layer) that regress heatmaps, one for each landmark. The latter has become popular, especially through hourglass-like architectures, which stack encoder-decoder networks with intermediate supervision to better capture spatial relationships~\cite{bulat2017far}. Landmark heatmaps can also be used to transfer information between stages during cascading regression using coordinate regression \cite{kowalski2017deep}. 

FLL does not necessarily have to be treated independently and can be learned together with correlated facial attributes using multi-task networks \cite{zhang2016learning}. It helps increase individual performance on each task. While most authors focus on the variance of faces, the intrinsic variance of image styles can also be handled to improve performance using style-aggregated networks \cite{dong2018san}.

\subsubsection{Localization in videos}

Recent work has shown that temporal coherence can be used to cope with facial and environmental variability under uncontrolled conditions. The most recent approaches generally combine convolutional neural networks (CNNs) and recurrent neural networks (RNNs) while decoupling the processing of spatial and temporal information to better leverage their complementarity \cite{liu2018two}. This late temporal connectivity helps stabilize predictions and handle global motion such as head pose variations. An unsupervised approach based on the coherency of optical flow can encourage temporal consistency in image-based detectors, which can reduce the jittering of landmarks in videos \cite{dong2018supervision}. The statistics of different kinds of movements can also be learned using a stabilization model designed to address time delays and smoothness issues \cite{tai-FHR-2019}. To go further, local motion can be included using early temporal connectivity based on 3D convolutions \cite{bel2018local}. By improving the temporal connectivity, more accurate predictions can be obtained, especially during expression variations. 

\subsubsection{3D / Multi-view Localization}

\romain{Another trend is the use of depth information to improve the accuracy of landmarks. The vast majority of approaches consider the face as a 2D object, which may cause issues in dealing with some transformations, e.g. out-of-plane rotations. 
A 3D Morphable Model (3DMM) can be fit to 2D facial images \cite{zhu2019face}. More recently\cite{bulat2017far}, 3D landmarks were also directly estimated from 2D facial images. In a multi-view context, epipolar constraints can be exploited \cite{dong2020tri}. 3D landmarks can be obtained through triangulation based on the predicted 2D landmarks from all views. After reprojection of the 3D landmarks, the difference between them and the 2D ones is computed to enforce multi-view consistency.}

\subsection{Scope of the paper}
\label{scope}

\benjamin{Although FLL approaches grow more and more robust, it is not clear today whether current performance is suitable for FER. We find it difficult, with the commonly used evaluation protocols, to understand the impact of FLL on FER.}

\benjamin{The aim of this work is on evaluating the aptitudes of various FLL approaches in the presence of head pose and facial expression and the suitability of the detected landmarks for FER. To this end, we have investigated several questions:}

\begin{itemize}[leftmargin=*]

\item \benjamin{\textit{How do current FLL approaches perform in the presence of pose and expression variations?}  In this work, we selected and evaluated the most representative and available approaches from the literature. Among the selected approaches are coordinate and heatmap deep regression models, including a multi-task one. In addition to quantifying the performance of these approaches according to the difficulties encountered in uncontrolled conditions, we studied their impact on FER. Note that, unfortunately, the temporal components of the recent dynamic approaches \cite{tai-FHR-2019,dong2018supervision} are not provided by the authors (at the time of writing). Moreover, 2D and 3D landmark predictions are not easily comparable. So, the focus has been narrowed down to a static 2D landmark evaluation. This is discussed in Section~\ref{landmark}.} 

\item \benjamin{\textit{What is the impact of FLL approaches on FER?} We used representative static and temporal approaches
to study the impact of FLL on FER. In this way, the recognition process is fully understood and experiments are facilitated. This issue is addressed in Section~\ref{expression}.}

\item \benjamin{\textit{Is the commonly used FLL evaluation metric relevant to FER? } By computing
correlation coefficients between FLL and FER performances, we show that the commonly used FLL evaluation metric is a poor predictor of FER performances. In the light of the conclusions drawn from Section~\ref{expression}, we investigated several new FLL evaluation metrics adapted to FER that account for stability over time and the importance of some specific landmarks. We addressed this issue in Section~\ref{metrics}.}

\end{itemize}

\benjamin{To answer these questions, it is important to define an evaluation protocol that covers both facial expression analysis and landmark detection in the presence and absence of head movements in video sequences. In recent years, many datasets have been proposed, which helped improve the robustness of FER and FLL approaches \cite{li2020deep, deng2018menpo}. Although these datasets include a large range of difficulties, they do not allow for the proper identification and quantification of the weaknesses of approaches according to each of these difficulties. They lack suitable contextual annotations and do not contain data captured in the absence of such difficulties, which is required to fully understand their impact. With the emergence of new datasets, such as SNaP-2DFe \cite{allaert2018impact}, data annotated with facial landmarks, facial expressions, and head movements are now available. In SNaP-2DFe, subjects are recorded both in the presence and absence of specific difficulties (i.e., head pose variations, facial expressions), along with contextual annotations and synchronized multi-camera capture. Other datasets containing contextual annotations include WFLW \cite{wu2018look}. However, they contain only still images and the contextual annotations; although they are numerous (i.e., expressions, poses, occlusions, make-up, illumination, and blur), they remain limited as no details beyond the category are provided. For these reasons, we use the SNaP-2DFe dataset in our experiments. This makes it possible to evaluate the robustness of recent FLL approaches according to these difficulties, and to study its impact on FER.}

\benjamin{Overall, we found that most approaches aim to optimize facial landmarks globally. Some approaches deal with related facial analysis tasks by relying on multi-task learning. Additional dimensions are also considered, including depth and temporal dimensions. As an example, stability over time is being considered carefully, probably due to the increase in the video content availability. However, all these works have objectives quite apart from subsequent tasks such as FER. There is, somewhat, the exception of multi-task approaches, but they do not address explicitly this issue. Is it the right choice? It is unclear today whether the current performance is sufficient for the purposes for which it is used.}

\benjamin{The problems faced by FLL approaches in localizing landmarks in presence of expressions and head pose variations have also a large impact on FER accuracy. We show that the localization errors of some landmarks have more impact on FER results.}

\benjamin{Considering the stability of landmarks over time and their importance, the new FLL evaluation metric proposed improves rank correlation consistently. This shows the benefit of considering FER specificities when measuring FLL performances.}

\section{Experimental protocol}
\label{dataset}

In this section we present the selected dataset, FLL approaches, FER approaches and evaluation metrics used in our experiments.

\subsection{Dataset}

\begin{figure}[h!]
\centering
\includegraphics[width=0.7\columnwidth]{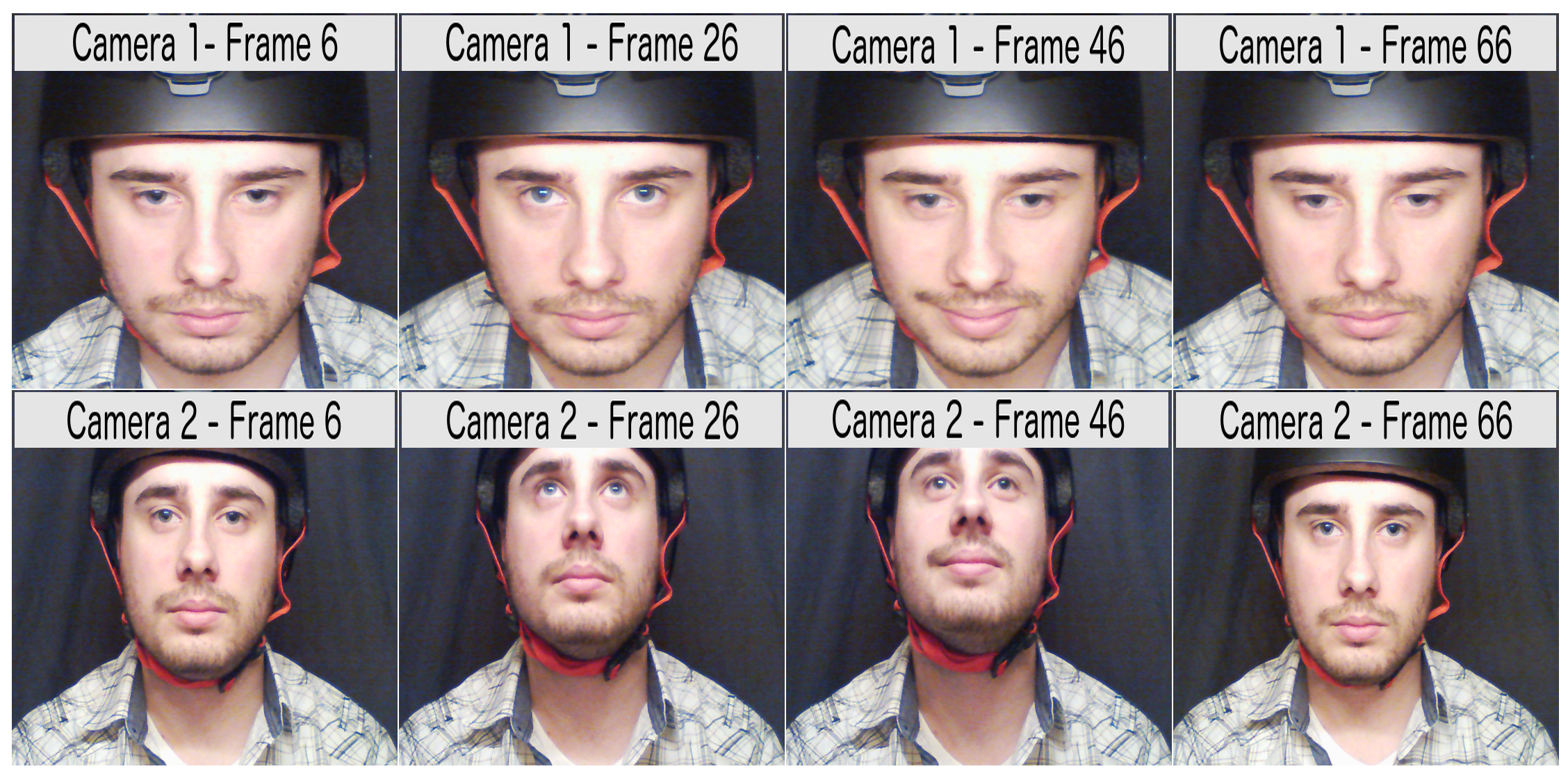}
\caption{Sample images of facial expressions recorded under pitch movements from the SNaP-2DFe dataset \cite{allaert2018impact} (row 1: helmet camera, expression only; row 2: static camera, both expression and head pose variations).}
\label{setting}
\end{figure}

Unlike other datasets, SNaP-2DFe \cite{allaert2018impact} has been collected simultaneously under constrained and unconstrained head poses, as illustrated in Figure \ref{setting}. Given its rich annotations, it represents a great candidate to conduct our study, allowing for a comprehensive analysis of both tasks regarding major difficulties.

SNaP-2DFe \cite{allaert2018impact} contains more than 90,000 images from 1,260 videos of 15 subjects. These videos contain image sequences of faces in frontal and non-frontal scenarios. For each subject, six head pose variations (static -- no head movement, translation-x -- along the x-axis, yaw, pitch, roll -- up to 60 degrees, and diagonal -- from the upper-left corner to the lower-right corner) combined with seven expressions (neutral, anger, disgust, fear, happiness, sadness, and surprise) were recorded by two synchronized cameras, resulting in a total of 630 constrained recordings (i.e., without head movements) and 630 unconstrained recordings (i.e., with head movements). SNaP-2DFe also provides annotations of the temporal patterns of expression activations (neutral-onset-apex-offset-neutral). Sixty-eight facial landmarks have been initially localized using the method in \cite{kazemi}. All frames were then individually inspected and, when needed, re-annotated in order to compensate for landmark localization errors.

\subsection{Facial landmark localization approaches}

Given the large number of facial localization approaches in the literature, we have selected a subset of recent approaches representative of the current state of the problem. We focus on deep learning-based approaches as they currently constitute the dominant trend. Among them, we selected state-of-the-art models for each of the categories highlighted in Section~\ref{section:review:alignment}:
\begin{itemize}
    \item coordinate regression models: Deep Alignment Network (DAN)~\cite{kowalski2017deep};
    \item heatmap regression models: HourGlass (HG)~\cite{bulat2017far} and Style Aggregated Network (SAN)~\cite{dong2018san};
    \item multi-task models: Tasks-Constrained Deep Convolutional Network (TCDCN)~\cite{zhang2016learning};
    \item dynamic models (without their temporal components\footnote{The code and models of the temporal components are not provided by the authors.}): Supervision By Registration (SBR)~\cite{dong2018supervision} and Fractional Heatmap Regression (FHR)~\cite{tai-FHR-2019}.
\end{itemize}

Since the code and the pre-trained models provided by the respective authors are used, it is important to consider the datasets used to train these approaches to ensure that there is no significant bias in our evaluation. As shown in Table \ref{FLBDD}, most approaches are trained mainly on 300W, HELEN, and AFLW. HG, TCDCN, SBR, and FHR make use of additional datasets, either to improve performance by limiting overfitting, or to provide the necessary multi-task components.

\begin{table}[h!]
\centering
\fontsize{5}{6}\selectfont
\caption{Datasets used to train the different approaches selected for our study.}
\begin{tabular}{|c|c|c||c|c|c|c|c|c|}
\hline 
\multicolumn{3}{|c||}{Datasets} & \multicolumn{6}{c|}{FLL approaches} \\
\hline 
Name & Type & Images & HG & TCDCN & DAN & FHR & SBR & SAN \\
\hline 
300W \cite{sagonas2016300} & Static & 600 & \ding{51} & \ding{51} & \ding{51} & \ding{51} & \ding{51} & \ding{51} \\
\cellcolor{gray!10}{HELEN \cite{le2012interactive}} & \cellcolor{gray!10}{Static} & \cellcolor{gray!10}{2,330} & \cellcolor{gray!10}{\ding{51}} & \cellcolor{gray!10}{\ding{51}} & \cellcolor{gray!10}{\ding{51}} & \cellcolor{gray!10}{\ding{51}} & \cellcolor{gray!10}{\ding{51}} & \cellcolor{gray!10}{\ding{51}} \\
AFLW \cite{koestinger2011annotated} & Static & 25,993 & \ding{51} & \ding{51} & \ding{51} & \ding{51} & \ding{51} & \ding{51} \\
\cellcolor{gray!10}{COFW \cite{burgos2013robust}} & \cellcolor{gray!10}{Static} & \cellcolor{gray!10}{1007} & \cellcolor{gray!10}{-} & \cellcolor{gray!10}{\ding{51}} & \cellcolor{gray!10}{-} & \cellcolor{gray!10}{-} & \cellcolor{gray!10}{-} & \cellcolor{gray!10}{-} \\
MAFL \cite{zhang2016learning} & Static & 20,000 & - & \ding{51} & - & - & - & - \\
\cellcolor{gray!10}{300W-LP \cite{zhu2016face}} & \cellcolor{gray!10}{Static} & \cellcolor{gray!10}{61,225} & \cellcolor{gray!10}{\ding{51}} & \cellcolor{gray!10}{-} & \cellcolor{gray!10}{-} & \cellcolor{gray!10}{-} & \cellcolor{gray!10}{-} & \cellcolor{gray!10}{-} \\
Menpo \cite{zafeiriou20173d} & Static & 10,993 & \ding{51} & - & \ding{51} & - & - & - \\
\cellcolor{gray!10}{300VW \cite{shen2015first}} & \cellcolor{gray!10}{Temporal} & \cellcolor{gray!10}{218,595} & \cellcolor{gray!10}{\ding{51}} & \cellcolor{gray!10}{-} & \cellcolor{gray!10}{-} & \cellcolor{gray!10}{-} & \cellcolor{gray!10}{-} & \cellcolor{gray!10}{-} \\
\hline
\end{tabular}
\label{FLBDD}
\end{table}

\subsection{Facial expression recognition approaches}
\label{proto:exp}

Based on the landmarks provided by FLL approaches, face registration is applied on the data recorded by the static camera in order to correct head pose variations and obtain frontal faces (see Figure \ref{fig:landmarks}, page \pageref{fig:landmarks}). We have chosen a recent 3D approach among the different face registration approaches used in the literature to deal with head pose \cite{hassner2015effective}. This approach has the advantage to preserve facial expressions.

\pierre{We have selected two typical features for FER: appearance-based (LBP~\cite{lbp}) and motion-based (LMP~\cite{allaert2018advanced}). We also include CovPool \cite{acharya2018covariance}, which is a static deep learning-based model. We did not consider landmark-free approaches such as ExpNet \cite{chang17expnet} in the main study as this method yields poorer results than the landmark-based one (see Appendix \ref{appendixa}). CovPool is based on a custom shallow model, the variant number 4, as stated in the original paper. It takes as input aligned face crops obtained through FLL. 
The SFEW 2.0 and RAF
datasets with standard augmentation techniques were used by the authors 
to train the model from scratch. The output of the 2nd fully connected layer (capacity of 512) serves as features.}

\pierre{For each method that we selected, feature extraction is performed only on the apex frame of each sequence. Apex frames correspond to the most favorable phase for FER and the most difficult one for FLL. We consider this phase as the most interesting for our study. Apex features are fed into an SVM classifier with an RBF kernel trained to recognize the 7 expressions.}


\section{Effectiveness of landmark localization}
\label{landmark}
Nowadays, it may be difficult to clearly understand the current state of the FLL problem. Hence, there is a need for benchmarks to better identify the benefits and limits of the various approaches proposed so far. Beyond overall performance, there is also a need to quantify the accuracy of these approaches according to the different difficulties. These measurements are especially relevant as FLL appears to be critical for FER, which is now shifting to uncontrolled conditions. To meet these needs, we investigate the robustness of the selected FLL approaches in the presence of head pose variations and facial expressions. An overall performance analysis is first performed in order to identify which problem is the most challenging. The analysis then focuses on each facial landmark individually in order to identify more precisely which facial regions are more difficult to characterize when confronted to these problems.


\subsection{\benjamin{Metric for facial landmark localization}}
\label{critere_landmark}

The mean Euclidean distance between the predicted landmarks and the ground truth (GT) normalized by the diagonal of the GT bounding box is used as evaluation metric as it is robust to pose variations \cite{chrysos2018comprehensive}. The error $e_n$ for image $n$ is expressed as:
\begin{equation}
e_n = \frac{1}{L} \overset{L}{\underset{i=1}{\sum}} \frac{||p_{i} - g_{i}||_2}{D}.
\label{equation:euclideanerror}
\end{equation}
\noindent where $L$ is the number of landmarks, $p_i$ is the coordinates of the $i$-th predicted landmark, $g_i$ is the coordinates of the corresponding ground truth landmark, and $D$ is the length of the diagonal of the ground truth bounding box ($D = \mathrm{round}\left(\sqrt{{w}^{2} + \mathrm{h}^{2}}\right)$, with \(w\) and \(h\) the width and height of the bounding box, respectively). From this metric, we compute the area under the curve (AUC) and the failure rate (FR) with threshold $\alpha = 0.04$. Above this threshold, we consider a prediction as a failure, since facial components are mostly not located correctly. The AUC and FR are expressed as: 
\begin{equation}
\mathrm{AUC}_{\alpha} = \overset{\alpha}{\underset{0}{\int}} f(e)de.
\end{equation}
\begin{equation}
\mathrm{FR}_{\alpha} = 1 - f(\alpha).
\label{equation:AUC}
\end{equation}
\noindent where $f$ is the cumulative error distribution (CED) function and $\alpha$ the threshold. 

\subsection{Overall performance analysis}

In this experiment, the objective is to determine which movement caused by head pose variations or facial expressions has the biggest impact on FLL. Table~\ref{landmarks_t2} shows the performance of the selected approaches in the presence of head pose variations only (i.e., neutral face with the 6 head movements), facial expressions only (i.e., static frontal face with the 6 facial expressions), and head pose variations combined with facial expressions (i.e., the 6 facial expressions along with the 6 head movements). For each approach, AUC and FR are computed over the entire sequences.

\begin{table}[!h]
\centering
\fontsize{4.3}{6}\selectfont
\caption{AUC$/$FR with and without head pose variations and facial expressions on SNaP-2DFe. Static means no expression and no head movement.}
\begin{tabular}{|c|c|c|c|c||c|}
\cline{2-6} \multicolumn{1}{c|}{} & Static & Pose variations only & Expressions only & Pose variations \& expressions & Average \\
\hline
HG & 56.57 / 0.00 & 54.30 / 0.42 & 55.35 / 0.05 & 52.94 / 0.58 & 54.79 / 0.26 \\
\hline 
\cellcolor{gray!10}{TCDCN} & \cellcolor{gray!10}{59.98 / 0.00} & \cellcolor{gray!10}{54.86 / 4.49} & \cellcolor{gray!10}{59.73 / 0.00} & \cellcolor{gray!10}{52.97 / 5.01} & \cellcolor{gray!10}{56.88 / 2.37} \\
\hline
DAN & 72.05 / 7.21 & 68.73 / 8.44 & 71.77 / 6.41 & 67.88 / 8.34 & 70.10 / 7.60 \\
\hline
\cellcolor{gray!10}{FHR} & \cellcolor{gray!10}{72.51 / 0.00} & \cellcolor{gray!10}{69.90 / 0.17} & \cellcolor{gray!10}{71.84 / 0.00} & \cellcolor{gray!10}{68.51 / 0.65} & \cellcolor{gray!10}{70.69 / 0.20} \\
\hline
SBR & 74.64 / 0.45 & 70.44 / 1.02 & 73.76 / 0.57 & 68.86 / 2.02 & 71.92 / 1.01 \\
\hline
\cellcolor{gray!10}{SAN} & \cellcolor{gray!10}{73.99 / 0.00} & \cellcolor{gray!10}{70.97 / 0.21} & \cellcolor{gray!10}{73.80 / 0.00} & \cellcolor{gray!10}{70.27 / 0.44} & \cellcolor{gray!10}{72.25 / 0.16} \\
\hline
\hline
Average & 68.29 / 1.27 & 64.87 / 2.46 & 67.71 / 1.17 & 63.57 / 2.84 & 66.10 / 1.93 \\
\hline
\end{tabular}
\label{landmarks_t2}
\end{table}

Table \ref{landmarks_t2} indicates that the AUC is lower and the FR higher in the presence of head pose variations than in the presence of facial expressions for all approaches. These results suggest that head pose variations are more challenging than facial expressions for FLL. The appearance is generally much more affected by a change in pose than a change in expression. The geometry of landmarks is affected by the pose, and it challenges the geometric priors of the model. 
In the simultaneous presence of both difficulties, the performance tends to decrease further.

In order to obtain more accurate performance measurements for each approach according to head pose variations and facial expressions, the previous results on the full set (i.e., head pose variations and expressions) are detailed by activation pattern in Table \ref{landmarks_t3}. Expressions and most poses are mainly active at the apex.

\begin{table}[!h]
\centering
\fontsize{5}{6}\selectfont
\caption{AUC$/$FR by activation pattern on SNaP-2DFe in the presence of facial expressions.}
\begin{tabular}{|c|c|c|c|c|c|}
\cline{2-6} \multicolumn{1}{c|}{} & Neutral to Onset & Onset to Apex & Apex to Offset & Offset to Neutral & All\\
\hline
HG & 55.21 / 0.06 & 49.30 / 1.73 & 48.58 / 1.56 & 51.74 / 0.65 & 52.94 / 0.58\\
\hline 
\cellcolor{gray!10}{TCDCN} & \cellcolor{gray!10}{55.55 / 3.17} & \cellcolor{gray!10}{44.70 / 11.77} & \cellcolor{gray!10}{45.99 / 9.64} & \cellcolor{gray!10}{51.84 / 4.92} & \cellcolor{gray!10}{52.97 / 5.01}\\
\hline
DAN & 69.87 / 7.91 & 62.35 / 9.95 & 64.07 / 8.99 & 67.51 / 8.34 & 67.88 / 8.34 \\
\hline
\cellcolor{gray!10}{FHR} & \cellcolor{gray!10}{71.57 / 0.04} & \cellcolor{gray!10}{64.14 / 2.37} & \cellcolor{gray!10}{63.24 / 1.67} & \cellcolor{gray!10}{67.15 / 0.62} & \cellcolor{gray!10}{68.52 / 0.64}\\
\hline
SBR & 71.74 / 1.04 & 62.22 / 5.12 & 63.75 / 3.42 & 67.35 / 2.83 & 68.86 / 2.02 \\
\hline
\cellcolor{gray!10}{SAN} & \cellcolor{gray!10}{72.79 / 0.00} & \cellcolor{gray!10}{65.65 / 1.22} & \cellcolor{gray!10}{66.41 / 1.08} & \cellcolor{gray!10}{68.86 / 0.56} & \cellcolor{gray!10}{70.27 / 0.44}\\
\hline
\hline
Average & 66.12 / 2.04 & 58.06 / 5.36 & 58.67 / 4.39 & 62.41 / 2.99 & 63.57 / 2.84\\
\hline
\end{tabular}
\label{landmarks_t3}
\end{table}

As shown in Table \ref{landmarks_t3}, the accuracy of all approaches decreases the most for images adjacent to the apex state. More specifically, it corresponds to the moment where the expression and most head pose variations are at their highest intensity. As soon as the face gets closer to a neutral expression and a frontal pose, the accuracy improves. These results suggest that facial expressions with head pose variations remain a major difficulty for FLL. They show that not only the presence of an expression, but also its intensity, impact FLL as deformations are proportional to the intensity.

\subsection{Robustness to head pose variations}

In this experiment, we place the emphasis on head pose variations, by highlighting which type of head pose presents the most difficulties. In Table \ref{landmarks_t4}, the results are split by head pose variations. AUC and FR are computed on all sequences with a neutral face between the onset and the offset phase. The results ($\Delta\mathrm{AUC}$ and $\Delta\mathrm{FR}$) correspond to the difference in performance between a frontal neutral face and a neutral face in the presence of the different head pose variations. Negative (resp. positive) values for $\Delta\mathrm{AUC}$ (resp. $\Delta\mathrm{FR}$) indicate difficulties in the presence of the given head pose variation.

\begin{table}[h!]
\centering
\fontsize{4}{6}\selectfont
\caption{$\Delta\mathrm{AUC}$ / $\Delta\mathrm{FR}$ for each head pose variation on SNaP-2DFe. Red cells indicate decreases in AUC.}
\begin{tabular}{|c|c|c|c|c|c||c|}
\cline{2-7} \multicolumn{1}{c|}{} & Translation-x & Roll & Yaw & Pitch & Diagonal & Average\\
\hline
HG & \cc{0}-0.89 / 0.0 & \ccc{0}+0.67 / 0.0 & \cc{4}-4.17 / 0.0 & \cc{1}-1.19 / +0.67 & \cc{10}-10.18 / +3.67 & \cc{3}-3.15 / +0.86\\
\hline 
TCDCN & \cc{0}-0.36 / 0.0 & \cc{1}-1.36 / 0.0 & \cc{21}-21.33 / +24.0 & \cc{9}-9.39 / 0.0 & \cc{30}-30.65 / +33.67 & \cc{12}-12.61 / +11.53\\
\hline
DAN  & \cc{4}-4.0 / +2.0 & \cc{13}-13.15 / +15.0 & \cc{1}-1.76 / -1.0 & \cc{8}-8.8 / +0.66 & \cc{15}-15.79 / +4.33 & \cc{8}-8.7 / +4.19\\
\hline
FHR & \cc{0}-0.97  /  0.0 & \cc{1}-1.69  /  0.0 & \cc{0}-0.66  /  0.0 & \cc{10}-10.08  /  0.0 & \cc{7}-7.5  /  +2.0 & \cc{4}-4.18 / +0.4 \\
\hline
SBR & \cc{4}-4.21 / -0.33 & \cc{2}-2.82 / -1.0 & \cc{7}-7.15 / -0.33 & \cc{12}-12.36 / +0.0 & \cc{16}-16.71 / +1.67 & \cc{8}-8.65 / 0.0\\
\hline
SAN & \cc{0}-0.75 / 0.0 & \cc{5}-5.22 / 0.0 & \cc{2}-2.36 / 0.0 & \cc{8}-8.02 / 0.0 & \cc{11}-11.08 / 0.0 & \cc{5}-5.48 / 0.0 \\
\hline
\hline
Average & \cc{1}-1.86 / +0.28 & \cc{3}-3.93 / +2.33 & \cc{6}-6.24 / +3.78 & \cc{8}-8.31 / +0.22 & \cc{15}-15.32 / +7.56 & \cc{7}-7.13 / +2.83 \\
\hline
\end{tabular}
\label{landmarks_t4}
\end{table}

According to the results in Table \ref{landmarks_t4}, some head pose variations seem more difficult to handle than others. Diagonal and pitch lead to a severe drop in the AUC and a considerable increase in the FR, suggesting that these head pose variations are the most challenging. They involve out-of-plane rotations, some of them on several rotation axes, which have a significant impact on the appearance of the face and the geometry of the landmarks. This decrease in performance may also be related to an insufficient amount of training data for these movements. There is a drop in the AUC for yaw, roll, and translation-x as well, but with a $\Delta\mathrm{FR}$ of almost zero~\footnote{The average $\Delta\mathrm{FR}$ for roll and yaw is due to a single outlier (respectively DAN and TCDCN).}. This shows a better ability of the approaches to manage these variations. The errors generated are fairly small and do not result in localization failures. Even for SAN, which has the best overall performance, the average $\Delta\mathrm{AUC}$ remains high with a loss of -5.48 with, however, an average $\Delta\mathrm{FR}$ of 0.

\subsection{Robustness to facial expressions}

In this experiment, we focus on facial expressions, by highlighting which types of expressions present the most difficulties. In Table \ref{landmarks_t5}, the results are split by facial expression. AUC and FR are computed on all sequences where the face is frontal and static from the onset to the offset of the expression. The results ($\Delta\mathrm{AUC}$ and $\Delta\mathrm{FR}$) denote the difference in performance between a frontal neutral face and a frontal face with the different facial expressions. Negative (resp. positive) values for $\Delta\mathrm{AUC}$ (resp. $\Delta\mathrm{FR}$) indicate difficulties in the presence of the given facial expression.

\begin{table}[h!]
\centering
\fontsize{3.8}{6}\selectfont
\caption{$\Delta\mathrm{AUC}$ / $\Delta{\mathrm{FR}}$ for each facial expression on SNaP-2DFe. Red (resp. green) cells indicate decreases (resp. increases) in AUC.}
\begin{tabular}{|c|c|c|c|c|c|c||c|}
\cline{2-8} \multicolumn{1}{c|}{} & Happiness & Anger & Disgust & Fear & Surprise & Sadness & Average \\
\hline
HG & \cc{4}-4.85 / +0.36 & \cc{4}-4.41 / 0.0 & \cc{8}-8.68 / 0.0 & \cc{1}-1.75 / 0.0 & \cc{1}-1.35 / 0.0 & \cc{3}-3.39 / 0.0 & \cc{4}-4.07 / 0.06 \\
\hline 
TCDCN & \cc{3}-3.12 / 0.0 & \ccc{1}+1.51 / 0.0 & \cc{8}-8.1 / 0.0 & \cc{0}-0.02 / 0.0 & \ccc{0}+0.55 / 0.0 & \cc{4}-4.57 / 0.0 & \cc{2}-2.29 / 0.0\\
\hline
DAN & \ccc{0}+0.48 / -4.30 & \cc{0}-0.86 / -0.22 & \cc{7}-7.04 / +0.74 & \cc{1}-1.29 / -2.16 & \cc{2}-1.21 / -0.12 & \cc{2}-3.11 / -2.67 & \cc{2}-2.17 / -1.45 \\
\hline
FHR & \cc{1}-1.44  /  0.0 & \cc{2}-2.28  /  0.0 & \cc{6}-6.62  /  0.0 & \cc{0}-0.96  /  0.0 & \ccc{0}+0.93  /  0.0 & \cc{4}-4.09  /  0.0 & \cc{2}-2.41 / 0.0 \\
\hline
SBR & \ccc{0}+0.04 / -0.64 & \cc{0}-0.37 / -1.00 & \cc{7}-7.27 / -0.63 & \cc{3}-3.49 / +2.01 & \cc{0}-0.37 / -1.00 & \cc{4}-4.03 / -0.38 & \cc{2}-2.58 / -0.27 \\
\hline
SAN & \ccc{0}+0.34 / 0.0 & +1.0 / 0.0 & \cc{7}-7.41 / 0.0 & \cc{1}-1.52 / 0.0 & \cc{1}-1.61 / 0.0 & \cc{4}-4.12 / 0.0 & \cc{2}-2.22 / 0.0\\
\hline
\hline
Average & \cc{1}-1.43 / -0.76 & \cc{0}-0.90 / -0.20 & \cc{7}-7.52 / +0.02 & \cc{1}-1.51 / -0.03 & \cc{0}-0.51 / -0.19 & \cc{3}-3.89 / -0.51 & \cc{2}-2.62 / -0.28\\
\hline
\end{tabular}
\label{landmarks_t5}
\end{table}

The results reported in Table \ref{landmarks_t5} show that some facial expressions seem more difficult to handle than others. Disgust and sadness lead to a significant drop in the AUC. These facial expressions involve complex mouth motions with significant changes in appearance, which may explain why the different approaches have more difficulties handling them. Besides, they also present a wider range of activation patterns and intensities, that may be less present in the datasets considered for training. The decrease in the AUC is less marked for happiness, anger, fear, and surprise, which seem to be handled better. As previously, a large value of the average $\Delta\mathrm{AUC}$, -2.22, is observed on the best selected approach in terms of overall performance (SAN).

 \subsection{Analysis of the landmarks}

\begin{figure}[h!]
\includegraphics[width=\columnwidth]{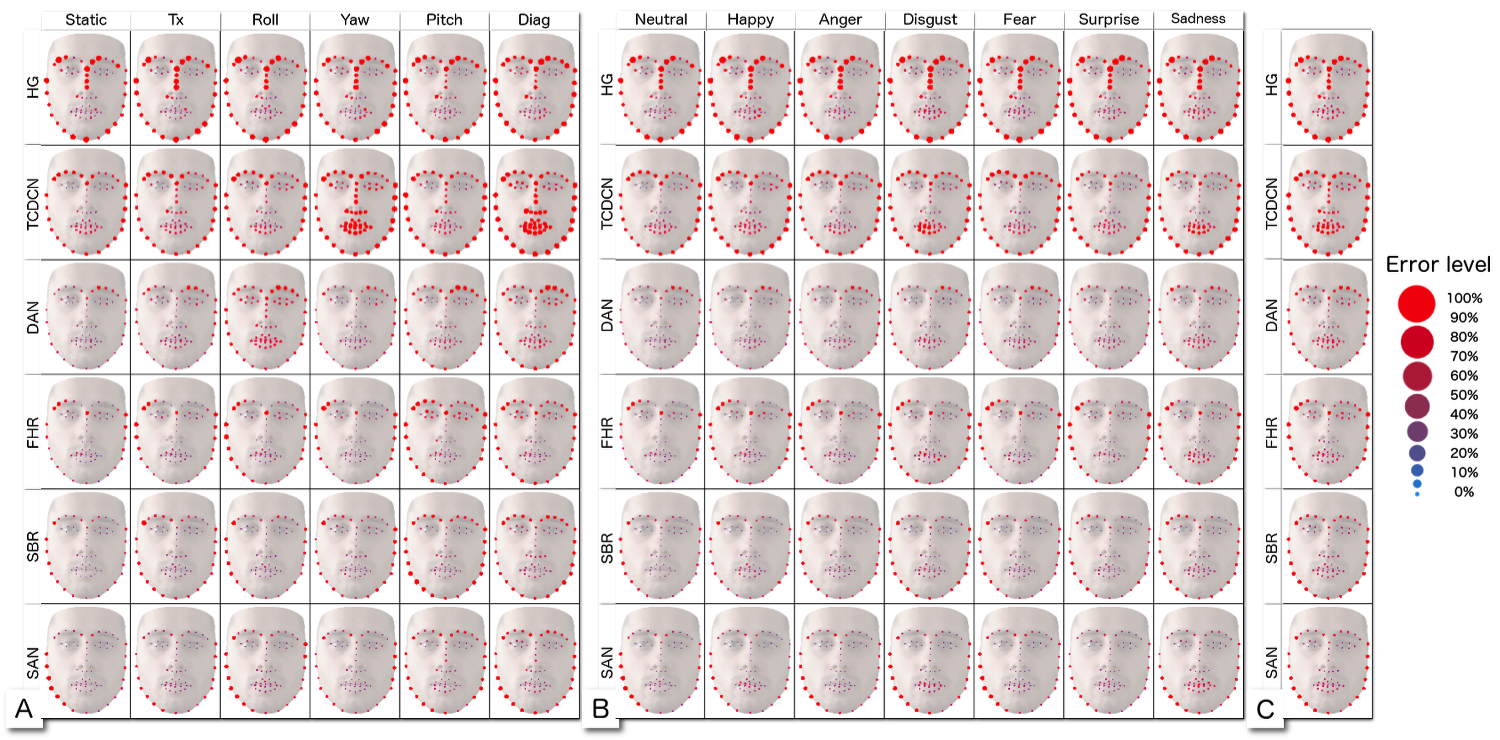}
\caption{Heatmaps of landmark localization error (A: per head pose, B: per facial expression, C: overall).}
\label{heatmap_error}
\end{figure}

To better identify the strengths and weaknesses of FLL approaches, a more detailed analysis at the level of each landmark is proposed. Figure \ref{heatmap_error} illustrates the landmark error levels of the different approaches according to each head pose variation (Figure \ref{heatmap_error}-A), each facial expression (Figure \ref{heatmap_error}-B) and the combination of both (Figure \ref{heatmap_error}-C). The landmark error levels are defined according to the threshold specified in Section \ref{critere_landmark}.

Regarding head pose variations (Figure \ref{heatmap_error}-A), in-plane transformations (static, translation-x, and roll) do not have a strong impact on most approaches. Eyebrows, the nose, and facial edges are among the regions where landmark localization lacks precision, especially for TCDCN, HG, and DAN. Out-of-plane (yaw, pitch and diagonal) variations, however, result in much more significant decreases of the accuracy. These transformations tend to stack up landmarks due to self-occlusions of certain parts of the face. Pitch and diagonal movements challenge all approaches. Eyebrows and facial edges are generally poorly detected. In the presence of facial expressions (Figure \ref{heatmap_error}- B), eyebrows and facial edges are also the most impacted regions for all approaches. Disgust and sadness are among the most challenging expressions as they produce complex inner motions, which also lead to poor localizations around the mouth. When both difficulties occur (Figure \ref{heatmap_error}-C), eyebrows, facial edges, and the mouth are therefore heavily affected. Note however that inner landmarks present less noise, especially around the mouth. This could simply be related to the training data (i.e., poor annotations of facial edges) or an implicitly stronger shape constraint during simultaneous variations.

\subsection{Discussion}

We show through these experiments that the accuracy of current FLL approaches is still impacted in presence of head pose variations and facial expressions. The combination of both, which is closer to the data captured under uncontrolled conditions, naturally increases the difficulty of the localization. The difficulties encountered are mainly related to certain expressions (disgust and sadness), due to their complex motion, and to certain head pose variations (pitch and diagonal), due to out-of-plane transformations. It seems very damaging that landmark localization is the weakest at the apex frame, which is the best phase for FER.

\benjamin{We have shown that eyebrows, facial edges, and the mouth are among the regions that are the most affected.
This may be explained by a poor annotation of these landmarks. It could also be related to a potential lack of distinctive features, especially for outline landmarks, and the propensity of landmarks to overlap in these regions. A possible solution could be to leverage the facial boundaries to which each point is strongly associated \cite{wu2018look}. These facial boundaries are easier to identify than inner landmarks in uncontrolled conditions and can help reduce ambiguity. So far, this approach is particularly interesting in the presence of occlusions, but less so in the presence of expressions.}

\benjamin{In addition to poor annotation, the limited availability of some conditions and unsuccessful data augmentation can be considered for further discussions. All these issues are more prominent for video data, most likely due its quantity and redundancy. Effective data augmentation can be crucial with current deep neural networks, especially when data under certain conditions is lacking. SAN, which relies on a smart form of augmentation, is a good example, as it is one of the most accurate approaches that have been evaluated. Hard sample mining strategies can help identify data that needs to be augmented \cite{feng2018wing}. However, constantly adding more data to improve performance is not always the most desirable solution, e.g., in terms of resources. Moreover, augmentation may be difficult or even impossible for extreme poses and expressions which are not necessarily found in the original data.} 

\romain{One way to alleviate these issues could be through recent advances in self-supervised learning and semi-supervised learning. Self-supervised aims to learn rich features that are transferable \cite{grill2020bootstrap}. It can be used as a pre-training step to improve supervised FLL \cite{he2020momentum}. Both labeled and unlabeled data can be used jointly to obtain abundant facial landmark labels, i.e., learning from partially labeled data \cite{dong2019teacher}. The ability to exploit large amounts of unlabeled data could help improve the robustness of FLL, especially for video, making us no longer be constrained by the quality and quantity of manual annotations. Besides, when a specific landmark scheme is not required, unsupervised FLL \cite{zhang2018unsupervised} could also be leveraged for FER instead of supervised FLL. However, given that the landmarks detected by these approaches are not controlled, the suitability of unsupervised FLL for subsequent tasks like FER has yet to be demonstrated.}

Overall, the difficulties studied in this work, head pose variations and facial expressions, involve generally coherent movements. Leveraging the dynamic nature of the face appears to be necessary in order to obtain more stable predictions over time and bring more robustness to these difficulties \cite{bel2018local}. Besides, many of the subsequent tasks, especially FER, have a dynamic nature. This makes it particularly relevant to consider temporal information  in order to be consistent with these tasks. However, temporal information, and more specifically facial dynamics, are currently under-exploited. This may be due to a lack of suitable data to train such dynamic models. Note that the ground truth of the few existing datasets may also lack stability. Current approaches do not make sufficient use of the possible synergy between motion, appearance, and shape. We believe that this is the key to achieve robust FLL in uncontrolled conditions.

Without studying the impact of FLL on subsequent tasks, it appears difficult to really understand how to improve current approaches in a useful way rather than in terms of Euclidean distance. Depending on the application, it may not be necessary to accurately detect all facial landmarks to properly characterize a face. For instance, landmarks around the mouth and the eyebrows seem to be more representative of some facial expressions than the landmarks located on the edges of the face. In the next sections, we conduct work in this direction regarding FER.

\section{Impact of landmark localization on facial expression recognition}
\label{expression}

After investigating the effectiveness of current FLL approaches, we look at their impact on FER. In this section, we evaluate how inaccurate FLL impacts FER through face registration. We consider two experimental settings. First, we train a classifier on frontal faces and then use registered faces for testing. Then, we train a classifier on registered faces and also use registered faces for testing. Hence, we can evaluate the impact of FLL in two common settings used for FER in presence of head pose variations. 


\subsection{\pierre{Evaluation protocol}}
Regarding FER, we apply a 10-fold cross validation protocol. For each evaluation, we report the average cross-validation accuracy. Stratification is performed both on facial expressions and head movements. When analyzing the results, one should be aware that face registration can sometimes be of poor quality (about 5 to 10\% of all images). This can result in reduced performances.

\subsection{Training based on frontal faces}

Data from the helmet camera (without head pose variations, see first row in Figure~\ref{setting}, page~\pageref{setting}) is used to train a facial expression classifier on frontal faces. Based on the predictions of the selected FLL approaches, we perform face registration on the data from the static camera (see second row in Figure \ref{setting}, page~\pageref{setting}) to obtain frontalized faces, which are used as the test set. Then, we evaluate the ability of FLL approaches to provide reliable inputs for face registration followed by FER. Different handcrafted and learned features (LBP \cite{lbp}, LMP \cite{allaert2018advanced}, CovPool \cite{acharya2018covariance}) are extracted to characterize facial expressions. Motion-based features (LMP) highlight the ability of FLL approaches to provide stable landmarks over successive images.

\begin{table}[h!]
\centering
\fontsize{4}{6}\selectfont
\caption{Comparison of FER performance according to different FLL approaches using texture-based (LBP and CovPool) and motion-based (LMP) features.}
\begin{tabular}{|c|c|c|c|c|c|c|c|c|c|}
\cline{2-10} \multicolumn{1}{c|}{} & \multicolumn{2}{c|}{Original face} & \multicolumn{7}{c|}{Registered face} \\
\hline
Feature & Helmet camera & Static camera & GT & HG & TCDCN & DAN & FHR & SBR & SAN \\
\hline LBP & 79.0 & 38.5 & 31.9 & 29.8 & 30.8 & 24.0 & 28.5 & 27.9 & 27.4 \\
\hline \cellcolor{gray!10}{LMP} & \cellcolor{gray!10}{84.2} & \cellcolor{gray!10}{31.3} & \cellcolor{gray!10}{33.7} & \cellcolor{gray!10}{24.3} & \cellcolor{gray!10}{24.7} & \cellcolor{gray!10}{22.1} & \cellcolor{gray!10}{14.0} & \cellcolor{gray!10}{25.9} & \cellcolor{gray!10}{31.8} \\
\hline
CovPool & 86.8 & 55.7 & 57.2 & 54.2 & 55.6 & 49.8 & 52.1 & 52.1 & 52.9 \\
\hline
\end{tabular}
\label{exp:tab1}
\end{table}

The results in Table \ref{exp:tab1} show that both texture-based and motion-based features give satisfactory performances in absence of head movements (i.e., on images from the helmet camera). However, a drastic fall in performance on the images from the static camera is observed due to head pose variations (e.g., roll, pitch, yaw) and large displacements (e.g., translation-x, diagonal). Note, however, the better ability of the deep learning approach (CovPool) to generalize, which is possibly due to its more diverse and discriminative features. Face registration based on the ground truth systematically leads to better FER accuracy, except for LBP. Still, it remains worse than the performances obtained on the data from the helmet camera. Most FLL approaches achieve performances that tend to be close to the ground truth. This is only true for texture-based features, not for motion-based features. The selected FLL approaches appear to be unstable over time, which is likely to result in temporal artifacts in registered face sequences. This tends to confirm the necessity for stable landmarks over time to drive face registration and keep the benefits of motion-based features. It represents a major obstacle to solving FER since motion-based recognition appears to be significantly better than texture-based recognition, as dynamic analysis has been shown to be more suitable for emotion recognition \cite{bassili1979emotion}. It is worth pointing out that HG and TCDCN, although among the weakest approaches in terms of FLL accuracy, perform similar to the best approaches when using texture-based features. Their lack of accuracy may be related to landmarks which are not critical for FER. Overall, these results show the importance of a suited protocol to ensure that face registration is beneficial. Note that training was performed on the data from the helmet camera, which does not contain registration artifacts. In the following, training is performed on the data from the static camera using face registration as a pre-processing step.

\subsection{Training based on registered faces}

In the following experiments, faces from the static camera are frontalized and used for both training and testing. We evaluate whether the FLL and face registration steps are able to preserve distinctive features for FER.

\subsubsection{Impact of head movements}

Table \ref{exp:tab2} presents the difference ($\Delta\mathrm{Acc}$) between the accuracy obtained using the ground truth landmarks and the accuracy obtained using the predictions of the selected FLL approaches during head movements. The raw scores using the ground truth landmarks are also provided to facilitate interpretation. 

\begin{table}[h!]
\centering
\fontsize{4.5}{5}\selectfont
\caption{$\Delta\mathrm{Acc}$ (\%) between registered faces based on the ground truth and the selected approaches during head movements. Static means no head movement. Red (resp. green) cells indicate decreases (resp. increases) in accuracy.}
\begin{tabular}{|c|c|c|c|c||c|c|c||c|}
\hline Land. & Feature & Static & Trans.-x & Roll & Yaw & Pitch & Diagonal & Average\\

\hline \multirow{3}{*}{GT} & LBP & 52.0 & 50.9 & 69.0 & 62.2 & 44.2 & 38.9 & 52.8 \\
& LMP & 69.5 & 61.8 & 61.1 & 65.4 & 56.7 & 56.9 & 61.9\\
& CovPool & 80.8 & 77.2 & 81.9 & 81.8 & 60.9 & 64.0 & 74.4 \\

\hline \multirow{3}{*}{HG} & LBP & \cc{0}-0.3 & \cc{10}-10.4 &  \cc{7}-7.0 & \cc{1}-1.2 & \cc{4}-4.7 & \cc{3}-3.0 & \cc{4}-4.4 \\
& LMP & \cc{12}-12.8 & \cc{17}-17.4 & \cc{17}-17.1 & \cc{16}-16.8 & \cc{17}-17.2 & \cc{9}-9.0 & \cc{15}-15.0 \\
& CovPool & \cc{4}-4.1 & \cc{1}-1.1 & \ccc{0}+0.9 & \cc{1}-1.0 & \cc{7}-7.2 & \cc{14}-14.1 & \cc{4}-4.4 \\

\hline \multirow{3}{*}{TCDCN} & LBP & \cc{11}-11.3 & \cc{9}-9.6 & \cc{16}-16.7 & \cc{7}-7.8 & \cc{8}-8.6 & \cc{9}-9.9 & \cc{10}-10.6 \\
& LMP & \cc{20}-20.4 & \cc{17}-17.7 & \cc{19}-19.8 & \cc{26}-26.8 & \cc{15}-15.6 & \cc{21}-21.0 & \cc{20}-20.2 \\
& CovPool & \cc{2}-2.6 & \ccc{1}+1.0 & \cc{1}-1.3 & \cc{4}-4.6 & \cc{0}-0.5 & \cc{4}-4.4 & \cc{2}-2.0\\

\hline \multirow{3}{*}{DAN} & LBP & \cc{7}-7.6 & \cc{9}-9.5 & \cc{17}-17.0 & \cc{9}-9.4 & \cc{3}-3.7 & \cc{14}-14.5 & \cc{10}-10.2\\
& LMP & \cc{16}-16.4 & \cc{15}-15.3 & \cc{20}-20.1 & \cc{22}-22.7 & \cc{17}-17.1 & \cc{17}-17.0 & \cc{18}-18.0\\
& CovPool & \cc{5}-5.7 & \cc{6}-6.1 & \cc{10}-10.2 & \cc{6}-6.4 & \cc{11}-11.8 & \cc{17}-17.6 & \cc{9}-9.6 \\

\hline \multirow{3}{*}{FHR} & LBP & \cc{2}-2.5 & \cc{0}-0.7 & \cc{11}-11.0 & \cc{6}-6.4 & \cc{13}-13.9 & \cc{14}-14.1 & \cc{8}-8.1\\
& LMP & \cc{9}-9.4 & \cc{11}-11.5 & \cc{2}-2.4 & \cc{12}-12.3 & \cc{9}-9.5 & \cc{1}-1.5 & \cc{7}-7.7\\
& CovPool & \ccc{0}+0.2 & \cc{1}-1.0 & \cc{3}-3.3 & +2.1 & \cc{14}-14.6 & \cc{13}-13.2 & \cc{4}-4.9\\

\hline \multirow{3}{*}{SBR} & LBP & +1.0 & \cc{1}-1.6 & \cc{11}-11.4 & \cc{1}-1.1 & \cc{10}-10.1 & \cc{10}-10.0 & \cc{5}-5.5\\
& LMP & \cc{6}-6.3 & \cc{13}-13.3 & \cc{5}-5.4 & \cc{21}-21.1 & \cc{15}-15.2 & \cc{18}-18.7 & \cc{13}-13.3\\
& CovPool & \cc{1}-1.8 & \cc{2}-2.2 & \cc{0}-0.3 & \cc{4}-4.0 & \cc{13}-13.4 & \cc{10}-10.2 & \cc{5}-5.3\\

\hline \multirow{3}{*}{SAN} & LBP & \cc{0}-0.7 & \cc{4}-4.6 & \cc{10}-10.0 & \cc{1}-1.8 & \cc{9}-9.4 & \cc{3}-3.4 & \cc{4}-4.9\\
& LMP & \cc{0}-0.8 & \cc{7}-7.1 & \cc{2}-2.1 & \cc{7}-7.0 & \cc{8}-8.0 & \cc{11}-11.1 & \cc{6}-6.0 \\
& CovPool & \ccc{0}+0.6 & \cc{0}-0.2 & \ccc{1}+1.7 & \cc{1}-1.6 & \cc{2}-2.8 & \cc{11}-11.7 & \cc{2}-2.3\\

\hline\hline \multirow{3}{*}{Average} & LBP & \cc{3}-3.5 & \cc{6}-6.0 & \cc{12}-12.1 & \cc{4}-4.6 & \cc{8}-8.4 & \cc{9}-9.1 & \cc{7}-7.2 \\
& LMP & \cc{11}-11.0 & \cc{13}-13.7 & \cc{11}-11.1 & \cc{17}-17.7 & \cc{13}-13.7 & \cc{13}-13.0 & \cc{13}-13.3 \\
& CovPool & \cc{2}-2.2 & \cc{1}-1.7 & \cc{2}-2.0 & \cc{2}-2.5 & \cc{8}-8.3 & \cc{11}-11.8 & \cc{4}-4.7\\
\hline
\end{tabular}
\label{exp:tab2}
\end{table}

Compared to Table \ref{exp:tab1}, scores are much higher with a training set based on registered data. However, it remains quite far (i.e., more than 10\%) from the scores obtained on data from the helmet camera. Regarding texture-based recognition, learned features (CovPool) give relatively small differences on in-plane transformations (static, translation-x, and roll), except for DAN. Handcrafted features (LBP), however, have difficulties during roll, for some approaches (i.e., HG, TCDCN, and DAN), and also during translation-x. Unsurprisingly, in the presence of out-of-plane transformations (yaw, pitch, and diagonal), larger differences can be observed than with in-plane transformations. Diagonal motion appears to be challenging for both features, as well as pitch, to a lesser extent. Overall, SAN provides the best performance, closely followed by HG. As a reminder, in our benchmark (Section~\ref{landmark}), SAN is the best FLL approach while HG is the worst.

As for motion-based recognition, handcrafted features (LMP) are strongly impacted by all movements. This is probably due to temporal artifacts. However, there is significantly better performance from SAN and FHR, which are likely to be more stable.

Note that some approaches (e.g., SAN) tend to perform better than the ground truth (e.g., static, roll). The face registration approach may have been trained on landmarks that are more similar to those provided by these approaches than those from the ground truth. It may also be due to an imperfect ground truth.

\subsubsection{Impact of landmark localization accuracy}

Table~\ref{exp:tab3} presents the difference ($\Delta\mathrm{Acc}$) between the accuracy obtained using the ground truth landmarks and the accuracy obtained using the predictions of the selected FLL approaches for each facial expression. In addition to the scores using the ground truth, scores on the helmet camera are also reported to provide insights about the intrinsic difficulty of the expressions.

\begin{table}[h!]
\centering
\fontsize{4.5}{5}\selectfont
\caption{$\Delta\mathrm{Acc}$ (\%) between registered faces based on the ground truth and the selected approaches for each facial expression. Red (resp. green) cells indicate decreases (resp. increases) in AUC.}
\begin{tabular}{|c|c|c|c|c|c|c|c|c|}
\hline Land. & Feature & Happi. & Fear & Surprise & Anger & Disgust & Sadness & Avg. \\

\hline \multirow{3}{*}{Helmet cam.} & LBP & 86.8 & 83.1 & 84.3 & 59.4 & 73.0 & 89.2 & 79.3\\
& LMP & 97.3 & 64.5 & 79.4 & 84.8 & 88.7 & 79.0 & 82.2 \\
& CovPool & 96.6 & 74.6 & 84.5 & 90.8 & 85.6 & 83.5 & 85.9 \\

\hline \hline \multirow{3}{*}{GT} & LBP & 65.3 & 54.7 & 60.3 & 30.7 & 47.4 & 54.4 & 52.1\\
& LMP & 80.5 & 29.4 & 54.8 & 66.3 & 58.4 & 57.8 & 57.8\\
& CovPool & 88.0 & 57.3 & 64.7 & 75.6 & 75.5 & 75.1 & 72.7\\

\hline \multirow{3}{*}{HG} & LBP & \cc{5}-5.3 & \cc{4}-4.9 & \cc{4}-4.9 & \ccc{0}+0.4 & \cc{1}-1.6 & \cc{10}-10.2 & \cc{4}-4.4  \\
& LMP & \cc{13}-13.7 & \ccc{1}+1.4 & \cc{16}-16.4 & \cc{18}-18.9 & \cc{15}-15.2 & \cc{14}-14.1 & \cc{12}-12.8 \\
& CovPool & \cc{2}-2.5 & \cc{3}-3.7 & \cc{3}-3.7 & \cc{5}-5.4 & \cc{4}-4.8 & \cc{8}-8.3 & \cc{4}-4.7 \\

\hline \multirow{3}{*}{TCDCN} & LBP & \cc{9}-9.6 & \cc{10}-10.0 & \cc{4}-4.3 & \cc{10}-10.7 & \cc{11}-11.4 & \cc{8}-8.4 & \cc{9}-9.0 \\
& LMP & \cc{23}-23.1 & \cc{9}-9.7 & \cc{17}-17.8 & \cc{31}-31.9 & \cc{12}-12.4 & \cc{20}-20.4 & \cc{19}-19.2 \\
& CovPool & \ccc{1}+1.7 & \cc{4}-4.7 & \cc{2}-2.5 & \cc{3}-3.0 & \ccc{0}+0.2 & \cc{2}-2.9 & \cc{1}-1.8\\

\hline \multirow{3}{*}{DAN} & LBP & \cc{14}-14.2 & \cc{12}-12.2 & \cc{7}-7.7 & \cc{7}-7.9 & \cc{10}-10.6 & \cc{2}-2.3 & \cc{9}-9.1\\
& LMP & \cc{27}-27.9 & \cc{4}-4.0 & \cc{17}-17.0 & \cc{17}-17.8 & \cc{9}-9.1 & \cc{21}-21.5 & \cc{16}-16.2\\
& CovPool & \cc{12}-12.2 & \cc{8}-8.7 & \cc{11}-11.3 & \cc{4}-4.6 & \cc{10}-10.5 & \cc{4}-4.7 & \cc{8}-8.6\\

\hline \multirow{3}{*}{FHR} & LBP & \cc{13}-13.0 & \cc{9}-9.7 & \cc{2}-2.9 & \cc{7}-7.2 & \cc{8}-8.1 & \cc{3}-3.7 & \cc{7}-7.4 \\
& LMP & \cc{11}-11.3 & \ccc{6}+6.0 & \cc{8}-8.2 & \cc{5}-5.3 & \cc{11}-11.7 & \cc{7}-7.0 & \cc{6}-6.2 \\
& CovPool & \cc{0}-0.6 & \cc{7}-7.6 & \cc{6}-6.3 & \cc{4}-4.2 & \cc{1}-1.3 & \cc{7}-7.9 & \cc{4}-4.6 \\

\hline \multirow{3}{*}{SBR} & LBP & \cc{10}-10.3 & \cc{3}-3.7 & \cc{7}-7.8 & \ccc{0}+0.4 & \cc{2}-2.0 & \cc{6}-6.2 & \cc{4}-4.9\\
& LMP & \cc{21}-21.8 & \cc{5}-5.9 & \cc{9}-9.4 & \cc{20}-20.3 & \cc{6}-6.6 & \cc{14}-14.0 & \cc{13}-13.0 \\
& CovPool & \cc{9}-9.4 & \cc{3}-3.1 & \cc{3}-3.1 & \cc{7}-7.0 & \cc{8}-8.3 & \cc{1}-1.1 & \cc{5}-5.3\\

\hline \multirow{3}{*}{SAN} & LBP & \cc{10}-10.3 & \cc{4}-4.2 & \cc{0}-0.7 & \ccc{1}+1.5 & \cc{7}-7.1 & \cc{2}-2.8 & \cc{3}-3.9 \\
& LMP & \cc{9}-9.7 & \ccc{4}+4.4 & \cc{9}-9.7 & \cc{11}-11.0 & \cc{4}-4.1 & \cc{3}-3.7 & \cc{5}-5.6\\
& CovPool & \cc{8}-8.0 & \cc{2}-2.2 & \cc{1}-1.6 & \cc{0}-0.2 & \cc{3}-3.2 & \ccc{2}+2.9 & \cc{2}-2.0 \\

\hline \multirow{3}{*}{Average} & LBP & \cc{10}-10.4 & \cc{7}-7.4 & \cc{4}-4.7 & \cc{3}-3.9 & \cc{6}-6.8 & \cc{5}-5.6 & \cc{6}-6.4\\
& LMP & \cc{17}-17.9 & \cc{1}-1.3 & \cc{13}-13.0 & \cc{17}-17.5 & \cc{9}-9.8 & \cc{13}-13.4 & \cc{12}-12.1\\
& CovPool & \cc{5}-5.1 & \cc{5}-5.0 & \cc{4}-4.7 & \cc{4}-4.0 & \cc{4}-4.6 & \cc{3}-3.6 & \cc{4}-4.5 \\
\hline
\end{tabular}
\label{exp:tab3}
\end{table}

LMP and CovPool are mostly suffering in presence of fear, while LBP is struggling with more expressions, especially anger. Happiness is one of the better handled expressions. Texture-based recognition results show smaller differences with learned features across all expressions. Handcrafted features have more difficulties with expressions such as happiness, fear, and disgust. Expressions involving significant deformations (e.g., of the lips) tend to have a significant impact on face registration. Overall, SAN and HG provides also the best performances.

The difference is once again larger for motion-based recognition (except for fear), probably because of unstable FLL, which introduces temporal artifacts and reduces performance. As for head movements, SAN and FHR perform better than the others.

\subsection{Impact per landmark}

\begin{figure}[h!]
\includegraphics[width=\columnwidth]{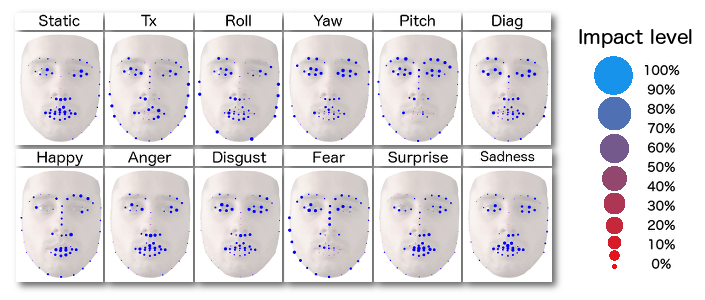}
\caption{Importance of each landmark to preserve facial geometry information according to head pose variations and facial expressions.}
\label{heatmap_impact}
\end{figure}

\benjamin{To study the impact of each of the 68 landmarks, we use linear regression and compute regression coefficients based on the FLL error of all approaches and the accuracy of FER. The importance of a landmark is evaluated according to the rank of its regression coefficient.} Figure \ref{heatmap_impact} shows the importance of facial landmarks to preserve facial geometry information according to head pose variations (first row) and facial expressions (second row).

In static frontal conditions, mainly landmarks on the lips are critical. During head pose variations, landmarks at the edges of the face and at the nasal ridge do not seem to be of major importance. The positions of the eyes, of the mouth, and of the center of the nose may be sufficient to correctly register the face. Eye areas appear to matter for out-of-plane transformations, to avoid deformations of the face. Note that for yaw and diagonal motions, the visible parts are as important as the occluded ones. We also observe that the landmarks at the edges of the face and at the nasal ridge are irrelevant when facial expressions occur. However, landmarks in the mouth area are essential for all expressions, except disgust in which eye areas are more critical.

\begin{figure}[h!]
\centering
\includegraphics[width=0.8\columnwidth]{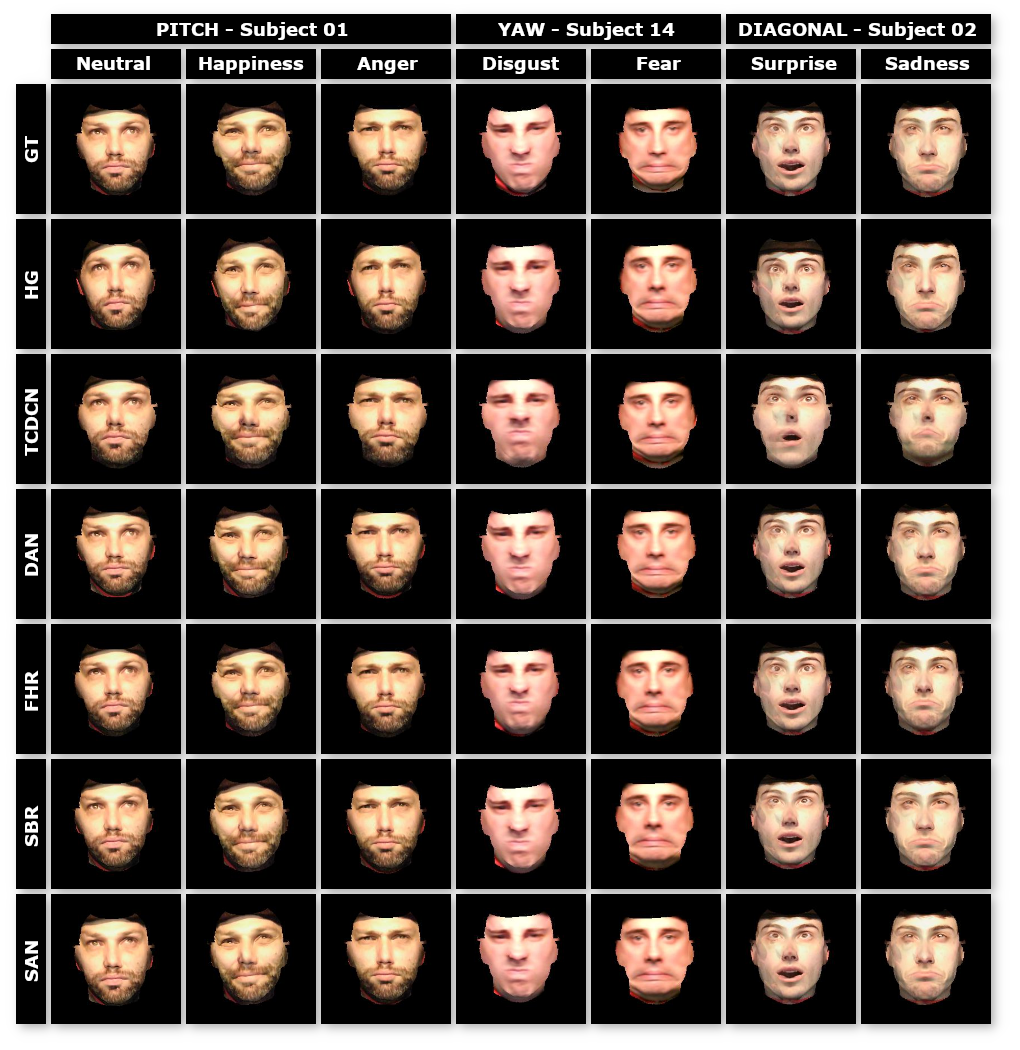}
\caption{Qualitative results of face registration.}
\label{ds2}
\end{figure}

Overall, landmarks located in the eye and mouth areas are essential to correctly register the face while maintaining its expression. Although landmarks at the nasal ridge appear to be irrelevant, note that landmarks at the bottom of the nose are still strongly present in all heatmaps. A possible explanation is that registration approaches, such as the one used in this work, take advantage of the pseudo-symmetry of the face. Figure \ref{ds2} illustrate the strong use of the pseudo-symmetry of the face during registration. Although this leads to unnatural faces, expressions are generally well preserved. Reconstruction artifacts can also be found. They are similar between approaches but may vary in intensity. This is probably related to the localization accuracy of the landmarks that are critical for face registration.

\subsection{Discussion}
In these experiments, we studied the ability of current FLL approaches to preserve facial geometry information during head pose variations and facial expressions for FER. Table \ref{exp:tab4} provides a summary of the results. HG and TCDCN perform poorly in terms of landmark precision, but can perform very well for FER. In contrast, FHR yields one of the best performances in terms of landmark precision, far ahead HG. However these results are not necessarily reflected on FER. Some facial regions are more important than others to characterize facial expressions, which could explain these results. It is more interesting to favor FLL approaches that correctly localize landmarks in the mouth and eye areas. Landmarks at the edges of the face were found to be irrelevant. Moreover, the $\Delta\mathrm{Acc}$ between the ground truth landmarks and the predictions of most of the selected approaches, when using motion-based features, is significantly larger than when using texture-based and deep learning-based features. This is related to the stability of landmark localization over time.

\begin{table}[h!]
\centering
\fontsize{5}{6}\selectfont
\caption{Summary of FLL performance in terms of AUC/FR (\%) and FER accuracy (\%) according to texture-based (LBP), motion-based features (LMP), and deep learning based features (CovPool).}
\begin{tabular}{|c|c|c|c|c|c|c|c|c|c|}
\cline{2-10}
 \multicolumn{1}{c|}{} & \multicolumn{3}{c|}{Landmark} & \multicolumn{2}{c|}{LBP} & \multicolumn{2}{c|}{LMP} & \multicolumn{2}{c|}{CovPool}\\ 
\cline{2-10}
 \multicolumn{1}{c|}{} & AUC & FR & Rank & Acc. & Rank & Acc. & Rank & Acc. & Rank\\
 \hline \cellcolor{gray!10}{GT} & \cellcolor{gray!10}{1.00} & \cellcolor{gray!10}{0.00} & \cellcolor{gray!10}{n/a} & \cellcolor{gray!10}{52.1} & \cellcolor{gray!10}{n/a} & \cellcolor{gray!10}{57.8} & \cellcolor{gray!10}{n/a} & \cellcolor{gray!10}{72.7} & \cellcolor{gray!10}{n/a} \\
 
 \hline

\hline HG & 55.35 & 0.05 & 6 & -4.4 & \textbf{2} & -12.8 & \textbf{3} & -4.7 & 4\\

\hline \cellcolor{gray!10}{TCDCN} & \cellcolor{gray!10}{59.73} & \cellcolor{gray!10}{0.00} & \cellcolor{gray!10}{5} & \cellcolor{gray!10}{-9.0} & \cellcolor{gray!10}{5} & \cellcolor{gray!10}{-19.2} & \cellcolor{gray!10}{6} & \cellcolor{gray!10}{-1.8} & \cellcolor{gray!10}{\textbf{1}} \\

\hline DAN & 71.77 & 6.41 & 4 & -9.1 & 6 & -16.2 & 5 & -8.6 & 6\\

\hline \cellcolor{gray!10}{FHR} & \cellcolor{gray!10}{71.84} & \cellcolor{gray!10}{0.00} & \cellcolor{gray!10}{\textbf{3}} & \cellcolor{gray!10}{-7.4} & \cellcolor{gray!10}{4} & \cellcolor{gray!10}{-6.2} & \cellcolor{gray!10}{\textbf{2}} & \cellcolor{gray!10}{-4.6} & \cellcolor{gray!10}{\textbf{3}} \\

\hline SBR & 73.76 & 0.57 & \textbf{2} & -4.9 & \textbf{3} & -13.0 & 4 & -5.3 & 5\\

\hline \cellcolor{gray!10}{SAN} & \cellcolor{gray!10}{73.80} & \cellcolor{gray!10}{0.00} & \cellcolor{gray!10}{\textbf{1}} & \cellcolor{gray!10}{-3.9} & \cellcolor{gray!10}{\textbf{1}} & \cellcolor{gray!10}{-5.6} & \cellcolor{gray!10}{\textbf{1}} & \cellcolor{gray!10}{-2.0} & \cellcolor{gray!10}{\textbf{2}} \\

\hline
\end{tabular}
\label{exp:tab4}
\end{table}

\pierre{The FER results obtained using FLL are still far from from those obtained with the ground truth landmarks, especially with motion-based features. Besides, the results obtained with the ground truth are not comparable to those obtained on the data from the helmet camera where no head movements is present (see Table \ref{exp:tab1} - column 1). This shows that there is still work to be done.}

These results highlight two main issues to be addressed to ensure that the subsequent tasks are taken into account and that FLL is improved towards these goals, which is currently not true.

\begin{itemize}
     
\item \benjamin{\textit{What landmarks should be used?} Some landmarks were found to be irrelevant for FER (i.e., landmarks at the edges of the face). As a result, in the context of FER, AUC and FR appear to be inappropriate. One solution would be to define application-specific landmark schemes that include only landmarks that are relevant to the targeted application. Another approach would be to customize FLL metrics to a given task by taking into account the relevance of each landmark to this task.}

\item \benjamin{\textit{What evaluation metric should be optimized?} The previous discussion questions the suitability of the evaluation metrics used for FLL in specific contexts. Besides landmark relevance, our experiments show that it is also necessary to better consider temporal information for FLL. Motion-based FER proves to be an effective approach but is also the one that is the most affected by landmark localization instability. More work should be done to take into account the temporal dimension when localizing landmarks, both in terms of evaluation metrics and models. This could greatly contribute to improve motion-based FER. Recently, a metric has been proposed to estimate the stability of an FLL approach \cite{tai-FHR-2019}. However, this metric relies on the assumption that the ground truth is stable, which is not always true. A few spatio-temporal models have also been proposed \cite{liu2018two}. They suffer from some limitations as well, e.g., they handle global movements better than local ones \cite{bel2018local}}.

\end{itemize}

\benjamin{In the following section, we offer some initial answers to these two questions by investigating new FLL evaluation metrics that cope better with FER.}


\section{\benjamin{Evaluation metrics revisited}}
\label{metrics}

\marius{We investigate whether the landmarks identified as impactful in our previous evaluations \pierre{and the temporal stability} can be used to improve the current FLL evaluation metrics. We aim to find a new FLL metric which is better correlated with FER performances.} \marius{We measure rank correlations since we analyze two variables (AUC and FER rate) that do not follow a normal distribution. In this case, the object of the study are not the raw values but their rank with respect to the other variable.} \pierre{Another reason for measuring rank correlation is practical: if FLL performance and FER accuracy are rank-correlated, then, when evaluating two FLL methods, one can predict, independently of their absolute performances, which one will perform best for FER.} \marius{Specifically, we could use Spearman's~\cite{spearman1961proof} and Kendall's~\cite{kendall1948rank} rank coefficients. These coefficients take values between -1 and +1. The closer they are to 1, the more positive the correlation (similar rankings of variables); the closer they are to -1, the more negative the correlation (reverse rankings). Finally, if the correlation score is close to zero, the probability that there is no monotonic link between the two variables is high.}

\marius{First, in the light of the conclusions of Section~\ref{expression}, we \pierre{extend the standard error metric of FLL (see Eq.~\ref{equation:euclideanerror}, page \pageref{equation:euclideanerror})} 
to also quantify the stability of landmarks over time. To do so, we consider the angle between the landmarks $p_{i,n}$ predicted at image $n$, the ground truth landmarks $g_{i,n}$ of the same image, and the landmarks $p_{i,n-1}$ of the previous image. This provides an error value between 0 and 1 corresponding to the angle that can vary from 0 to 180 degrees. The temporal angle $\beta_{i,n}$ for the $n$-th image and the $i$-th landmark is expressed as:}


\begin{equation}
\beta_{i,n} = \frac{\arccos \left(\frac{\langle\overrightarrow{g_{i,n}p_{i,n}},\overrightarrow{g_{i,n}p_{i,n-1}}\rangle}{||\overrightarrow{g_{i,n}p_{i,n}}||_2.||\overrightarrow{g_{i,n}p_{i,n-1}}||_2}\right)}{\pi}.
\end{equation}

\noindent \benjamin{where $g_{i,n}$ are the coordinates of landmark \(i\) of frame \(n\) in the ground truth, $p_{i,n}$ are the coordinates of predicted landmark \(i\) of frame \(n\) and \(\langle .,.\rangle\) is the inner product. The error $\epsilon_n$ for the $n$-th frame is then expressed as:}

\begin{equation}
\epsilon_n = \frac{1}{L} \overset{L}{\underset{i=1}{\sum}} (\beta_{i,n}+1)\cdot\frac{||p_{i,n} - g_{i,n}||_2}{D}.
\end{equation}

\noindent \benjamin{where $L$ is the number of landmarks, $p_{i,n}$ is the coordinates of the $i$-th predicted landmark in frame \(n\), $g_{i,n}$ is the coordinates of the corresponding ground truth landmark, and $D$ is the length of the diagonal of the ground truth bounding box ($D = \mathrm{round}\left(\sqrt{{w}^{2} + \mathrm{h}^{2}}\right)$, with \(w\) and \(h\) the width and height of the bounding box, respectively).} \pierre{We compare this metric to the standard Euclidean distance and to the distance-based stability measure proposed in~\cite{tai-FHR-2019}. For all three metrics, we measure FLL performance as the AUC (see Eq.~\ref{equation:AUC}, page~\pageref{equation:AUC}) with \(\alpha = 0.04\).}

\begin{figure}[h!]
\centering
\includegraphics[width=\columnwidth]{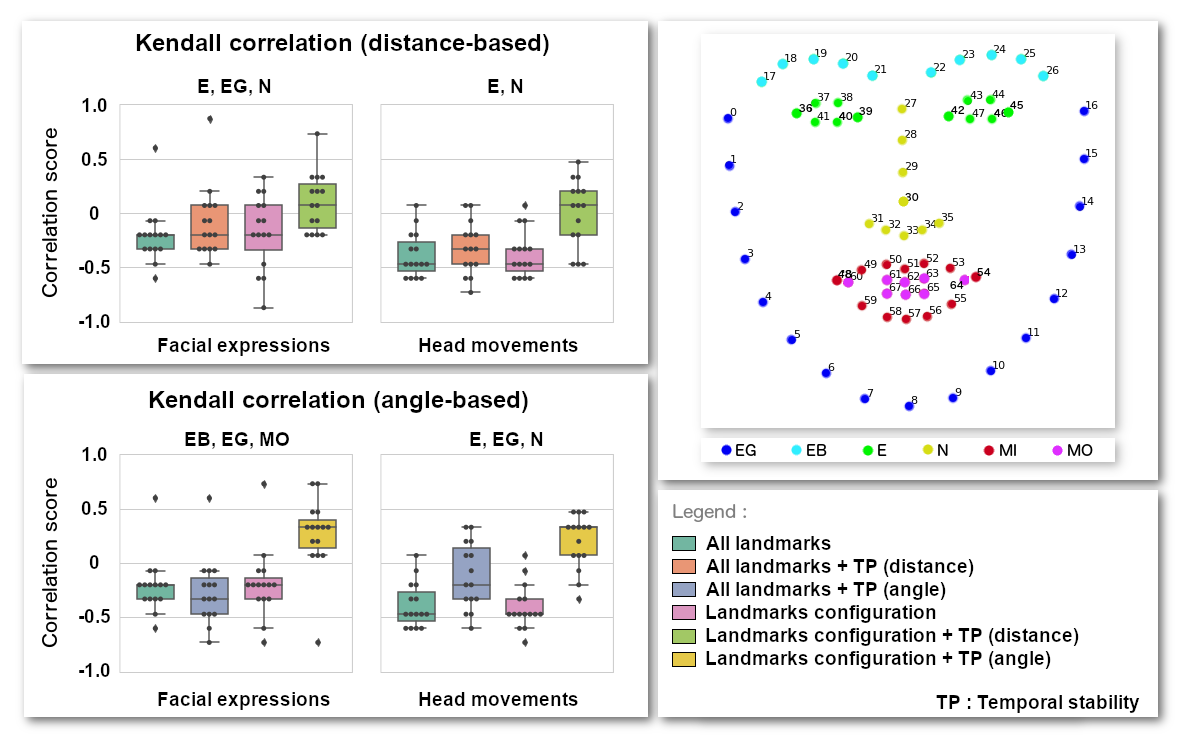}
\caption{\benjamin{Selected landmark schemes and correlation between the AUC of the different FLL metrics and FER accuracy, in the presence and absence of head pose variations.}}
\label{correl}
\end{figure}

\pierre{Then, we define the optimal subsets of landmarks to be considered in evaluating FLL in the context of FER, in two settings: in the presence and in the absence of head movements. To do so, we divide the standard 68-point landmark scheme into six subsets corresponding to different areas of the face (see Figure~\ref{correl}): edges (EG), eyebrows (EB), eyes (E), nose (N), inner mouth (MI), outer mouth (MO). We choose to analyze landmark subsets that target subregions of the face rather than working on a per-landmark basis. This allows us to have a more global view on the behaviors of the metrics. We evaluate the correlation between AUC and FER accuracy for all subsets and combinations of subsets of landmarks and each of the stability-based FLL metrics introduced previously (see Appendix~\ref{appendixb} for detailed experimental results). Finally, we select the landmark schemes that maximize this correlation in each setting and for each of these two metrics. It is important to note that it is difficult to identify an ideal configuration that adapts to all expressions and all movements. Besides, correlation values can vary significantly from one subject to another.}

\pierre{Figure \ref{correl} shows the selected landmark schemes and the correlations measured between the AUC of all FLL metrics and FER accuracy, for each subject of the SNaP-2DFe dataset, w.r.t. to each of the selected schemes and metrics. We report the values of Kendall's correlation coefficient only as it appears to be more stable than Spearman's coefficient in practice; the complete results with both correlation measures are provided in Appendix~\ref{appendixb}. Correlations are measured in two settings: in the absence of head movements (based on FER accuracy values from Table~\ref{exp:tab3}) and in their presence (based on FER accuracy values from Table~\ref{exp:tab2}).}


\pierre{The results show that the standard error metric for FLL and the distance-based stability metric are both little or negatively correlated with FER accuracy. The proposed metric based on space-time stability in terms of angle increases the correlation in a positive way, especially in the presence of head movements, since this is a situation where stable landmarks are the most needed to register the face correctly. Using task-specific landmark schemes provides a more significant improvement on the correlation, especially for the stability-based metrics. Still; the best correlation measured lies generally between 0.3 and 0.5.}

\pierre{The individual correlation scores of the subjects show a variety of situations, with some subjects showing high correlations between FLL performance and FER accuracy, while for some others they are much lower than the median. In the absence of head movements, we notice that subjects S1 and S5 tend to yield high correlation scores; these two subjects have the particularity to produce strong facial deformations when they display expressions. Subjects S9 and S12 tend to yield lower correlation scores; from their videos, we observe that their expressions have very low intensity. In the presence of head movements, subjects S6, S9, S11, and S12 yield higher correlation scores, while the scores drop for subjects S1, S2 and S15. This can be explained by the conditions of video acquisition, as the faces of some subjects were tilted with respect to the camera.}

\marius{In conclusion, we showed that changes in the FLL evaluation metric that account for temporal stability and landmark importance can lead to improved} 
\pierre{interpretation of FER performance from FLL performance, although the correlation scores could still be further improved.}


\section{Conclusion}
\label{conclusion}

The problem of FLL is currently being studied as an isolated problem when in fact it is a key component of many applications such as FER. Although performance has improved considerably with respect to the common metrics such as the Euclidean distance, we have shown that it does not guarantee that an approach with a lower mean error, a higher AUC, or a lower FR would give better performances when used for a subsequent task.

Another part of this study aimed at quantifying the impact of head poses and facial expressions on current FLL approaches. In the presence of specific head poses and facial expressions, the performance decreases significantly. In the third part, the impact of FLL on FER has been investigated. The results obtained from the ground truth landmarks are still far superior to those obtained from the predictions of recent FLL approaches. This margin is even more important when we compare these results to those obtained using the ground truth frontal faces. Due to the instability of landmark predictions over time, temporal approaches are more impacted. It is also clear that not all landmarks are equally relevant for FER, which further discredits the popular 68-points landmark scheme and metrics used to evaluate current FLL approaches. This is particularly noticeable with some approaches that provide among the worst results in terms of landmark localization, but perform among the best for FER.

\benjamin{From this study, two issues emerge: (1) Which landmarks should be used? (2) Which metric should be optimized? The Euclidean distance, as used in the literature, does not fit subsequent tasks such as FER. The new FLL evaluation metrics proposed in Section 6, which account for the specificity of FER, offer some answers to these questions.}

Suitable temporal modeling for FLL was also found to be of major importance to effectively capture the dynamics of the face. This complex dynamics involves both global motion, as in head movements, and local motion at the level of each components of the face. Such approaches are currently receiving little attention and much work remains to be done. It would also be interesting to study new loss functions based on the observations of this study to improve and adapt FLL for FER.

It is also desirable to find a suited way to integrate temporal and 3D approaches into this benchmark. SNaP-2DFe could be extended to other major difficulties, including occlusions. It is possible to artificially add static and dynamic occlusions to current data. The impact of dynamic occlusions is an interesting challenge for temporal approaches. Finally, beyond SNaP-2DFe, other datasets should be developed to evaluate FLL in other contexts.




%





\ifCLASSOPTIONcaptionsoff
  \newpage
\fi



%


\bibliographystyle{plain}
\bibliography{main}

%

\vspace{-1 cm}

\begin{IEEEbiographynophoto}
{Romain Belmonte}
received his MS degree (2015) and his Ph.D. (2019) in Computer Science from the University of Lille. His current research interests include computer vision, especially deep learning and human behavior analysis.
\end{IEEEbiographynophoto}

\vspace{-1 cm}

\begin{IEEEbiography}
[{\includegraphics[width=1in,height=1.25in,clip,keepaspectratio]{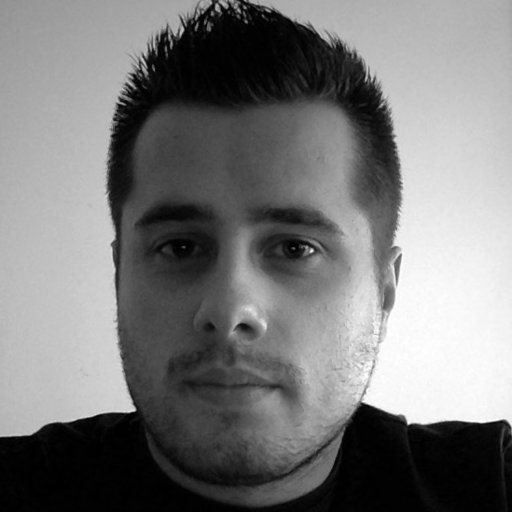}}]
{Benjamin Allaert}
received his MS degree on Image,Vision and Interaction and his Ph.D. on analysis of facial expressions in video flows in Computer Science from the University of Lille (France, 2018). He is currently a research engineer at the Computer Science Laboratory in Lille (CRIStAL). His research interests include computer vision and affective computing, and current focus of interest is the automatic analysis of human behavior.
\end{IEEEbiography}

\vspace{-1 cm}

\begin{IEEEbiographynophoto}
{Pierre Tirilly}
holds a M.Eng. in computer science from the Institut National des Sciences Appliqu\'ees (INSA, Rennes, France, 2006), a M.Sc. in Computer Science from INSA and Universit\'e de Rennes 1 (France, 2006), and a Ph.D. in Computer Science from Universit\'e de Rennes 1 (2010). From 2010 to 2012, he was a post-doctoral research fellow at University of Wisconsin-Milwaukee (USA). Since september 2012, he is an Assistant Professor at Universit\'e de Lille (France) and a member of the CRIStAL laboratory. His research interests include computer vision, multimedia, and bio-inspired computing.
\end{IEEEbiographynophoto}

\vspace{-1 cm}

\begin{IEEEbiography}
[{\includegraphics[width=1in,height=1.25in,clip,keepaspectratio]{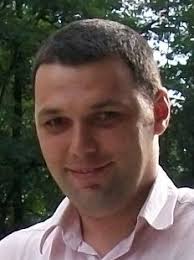}}]
{Ioan Marius Bilasco}
is an Assistant Professor at the University of Lille, France, since 2009. He received his MS degree on multimedia adaptation and his Ph.D. on semantic adaptation of 3D data in Computer Science from the University Joseph Fourier in Grenoble. In 2008, he integrated the Computer Science Laboratory in Lille (CRIStAL, formerly LIFL) as an expert in metadata modeling activities. Since, he extended his research to facial expressions and human behavior analysis.
\end{IEEEbiography}

\vspace{-1 cm}

\begin{IEEEbiography}
[{\includegraphics[width=1in,height=1.25in,clip,keepaspectratio]{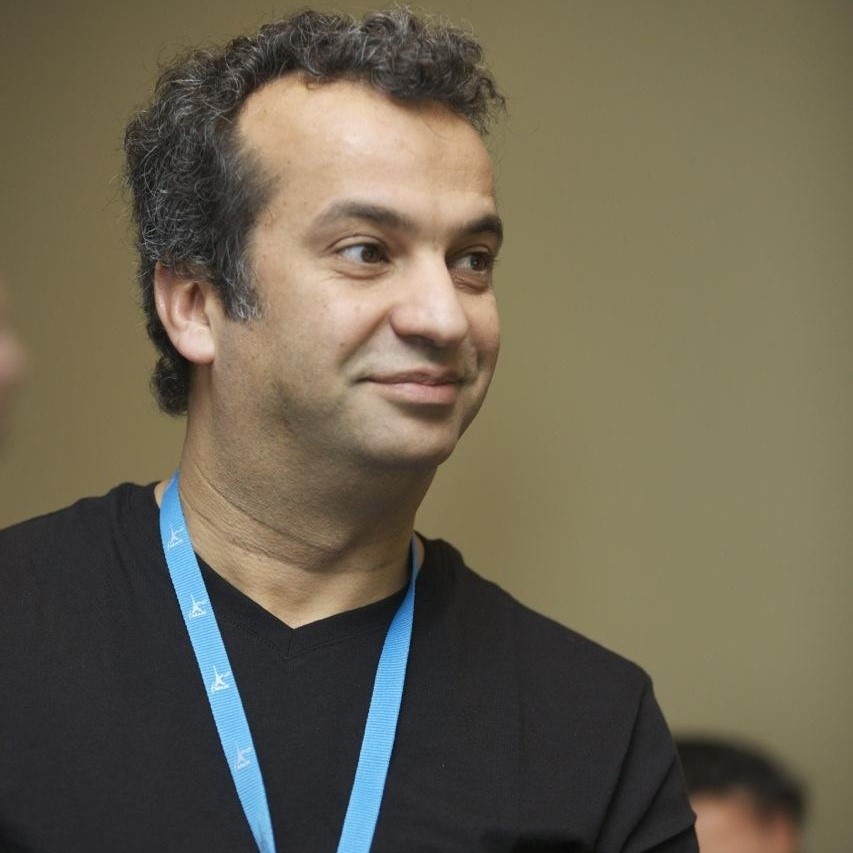}}]
{Chaabane Djeraba}
Chaabane Djeraba is full Professor at the University of Lille, France. He founded and leaded research group in the domain of multimedia, indexing and mining, with applications in human behavior understanding. He organized several special issues in journals. He chaired and founded several international workshops and sepcial sessions such as the international workshop on Content-Based Multimedia Indexing and Retrieval (CBMI) and ACM international workshop on Multimedia Data Mining. He took part in several program committees of conferences (e.g., ACM SIGKDD, ACM MM, IEEE MM).
\end{IEEEbiography}

\vspace{-1 cm}

\begin{IEEEbiography}
[{\includegraphics[width=1in,height=1.25in,clip,keepaspectratio]{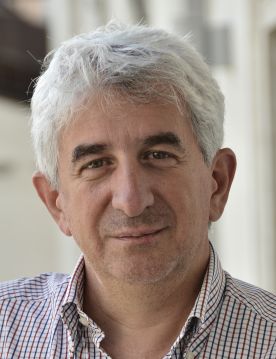}}]
{Nicu Sebe}
is a professor with the University of Trento, Italy, leading the research in the areas of multimedia information retrieval and human behavior understanding. He was the general co-chair of the IEEE FG 2008 and ACM MM 2013, and the program chair of the International Conference on Image and Video Retrieval in 2007 and 2010, ACM MM 2007 and 2011. He was the program chair of ICCV 2017 and ECCV 2016, and a general chair of ACM ICMR 2017. He is a fellow of the IAPR, and a senior member of the IEEE.
\end{IEEEbiography}

\appendices
\section{}
\label{appendixa}

\benjamin{In this section, we provide empirical evidence of the importance of facial landmarks for facial expression recognition. To this end, we have selected and evaluated state-of-the-art landmark-free and landmark-based approaches for FER, ExpNet \cite{chang17expnet} and CovPool \cite{acharya2018covariance}, respectively. Note that an extensive analysis of face registration for FER is already available in \cite{allaert2018impact}.}

\subsection*{\benjamin{Comparison between ExpNet and CovPool}}

\benjamin{ExpNet \cite{chang17expnet} and CovPool \cite{acharya2018covariance} are evaluated on original faces from the static camera (i.e., with head movements) under nominal conditions. \marius{CovPool is based on a custom shallow model, the variant number 4, as stated in the original paper. It takes as input aligned face crops obtained through FLL from data initially captured under uncontrolled conditions. ExpNet extracts 29 3DMM expression coefficients, which serve as features. ExpNet uses the ResNet-101 model trained from scratch using the raw face crops from the CASIA dataset~\cite{yi2014learning}, without augmentation.} ExpNet and CovPool models are not trained on the same training data.  ExpNet has obviously been trained under better conditions, with over 10 times more data.}

\begin{table}[!hbtp]
\centering
\fontsize{3.8}{7}\selectfont
\caption{\benjamin{Comparison of FER performance between landmark-free (ExpNet) and landmark-based (CovPool) approaches. The accuracies (\%) with original faces from the static camera and with registered faces based on the ground truth landmarks are provided. Original (0ri) means plain image for ExpNet and normalized faces for CovPool. Registered (Reg) means frontalized faces for both approaches. Static means no head movement.}}
\begin{tabular}{ccc||cccccccccc}
 & \multicolumn{2}{c}{All} & \multicolumn{2}{c}{Static} & \multicolumn{2}{c}{Roll} & \multicolumn{2}{c}{Yaw} & \multicolumn{2}{c}{Pitch} & \multicolumn{2}{c}{Diagonal}\\
 \cmidrule(l){2-3} \cmidrule(l){4-5} \cmidrule(l){6-7} \cmidrule(l){8-9} \cmidrule(l){10-11} \cmidrule(l){12-13}
 & Ori & Reg & Ori & Reg & Ori & Reg & Ori & Reg & Ori & Reg & Ori & Reg \\
 
\hline ExpNet & 55.8 & 69.0 & 68.0 & 74.1 & 56.0 & 78.6 & 67.1 & 75.7 & 38.4 & 52.7 & 32.3 & 53.3 \\

\cellcolor{gray!10}{CovPool} & \cellcolor{gray!10}{72.2} & \cellcolor{gray!10}{74.4} & \cellcolor{gray!10}{75.2} & \cellcolor{gray!10}{80.8} & \cellcolor{gray!10}{81.1} & \cellcolor{gray!10}{81.9} & \cellcolor{gray!10}{80.8} & \cellcolor{gray!10}{81.8} & \cellcolor{gray!10}{61.6} & \cellcolor{gray!10}{60.9} & \cellcolor{gray!10}{56.8} & \cellcolor{gray!10}{64.0} \\
\hline
\end{tabular}
\label{exp:deep}
\end{table} 

\benjamin{The results of the evaluation are provided in Table \ref{exp:deep}. Despite the possible bias related to the training data, overall, the landmark-based approach (CovPool) outperforms the landmark-free approach (ExpNet) by a large margin. Even when ExpNet integrates the registration step, its results remain below those obtained by CovPool on original faces. Face registration leads to a significant gain in performance for both models with a narrowing gap between them. There is an increase in accuracy of about 23.7\% for ExpNet and 3.1\% for CovPool. This tends to illustrate the benefits of face alignment and the difficulty that ExpNet has coping with non-frontal faces. We can observe a difference of about 33.3\% between the performances of ExpNet on original faces and the performances of CovPool with the complete recognition process studied in this work, i.e., with a face registration step. Results by head movement on original images show that CovPool also outperforms ExpNet for all movements. ExpNet is particularly struggling with pitch and diagonal movements. Face registration leads to a positive impact on CovPool for all movements, except for pitch, where the scores are close to each other. Results are even better for ExpNet where the difference between original and registered faces is more significant.}

\subsection*{\benjamin{Discussion}}

\benjamin{These results show the usefulness of FLL for FER under uncontrolled conditions. The common facial expression classification approach based on registered faces appears to perform significantly better than the facial expression regression approach. It is also interesting to see an improvement of ExpNet when using facial landmarks and normalization. This suggests that facial expression regression approaches also benefit from these preprocessings.}

\benjamin{The use of more supervision in the recognition process is one reason that may explain these results. Explicitly exploiting facial structure information is likely to be advantageous for a facial analysis task. Given these findings, can we consider that landmark-based approaches will always be better than landmark-free approaches as they use more supervision in the recognition process? Is this a problem in any way? Landmark annotations are available in most datasets. Besides, recent advances on unsupervised FLL \cite{zhang2018unsupervised} could be leveraged instead of supervised ones. In the end, is it a good idea for landmark-free approaches to avoid using landmarks? Some work suggests that the question is worth asking \cite{koujan2020real}.}

\benjamin{Be that as it may, we believe that facial landmarks should be used since they easily provide additional anchors for FER. A solution that includes landmark information has every chance to outperform a more naive solution. However, the accuracy of FLL plays an important role in the recognition process, as we demonstrated in the rest of the paper.}

\section{}
\label{appendixb}

\begin{figure}[h!]
\centering
\includegraphics[width=0.9\columnwidth]{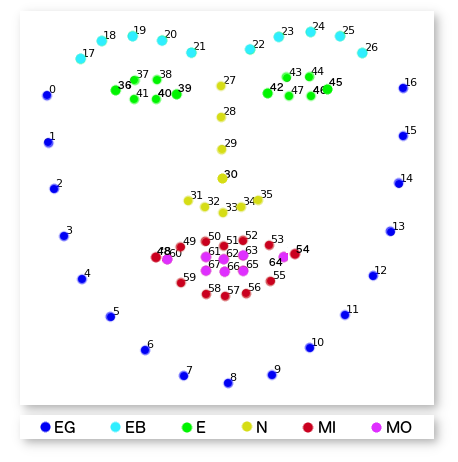}
\caption{\benjamin{Various landmark subsets covering specific facial regions: EG - edges, EB - eyebrows, E - eyes, N - nose, MI - inner mouth, MI - outer mouth, ALL - all landmarks.}}
\label{sets}
\end{figure}

\benjamin{In this appendix, we provide detailed results obtained while considering various subsets of landmarks covering simultaneously different parts of the face. The subsets covering specific facial regions are illustrated in Figure~\ref{sets}.}

\begin{figure}[h!]
\centering
\includegraphics[width=\columnwidth]{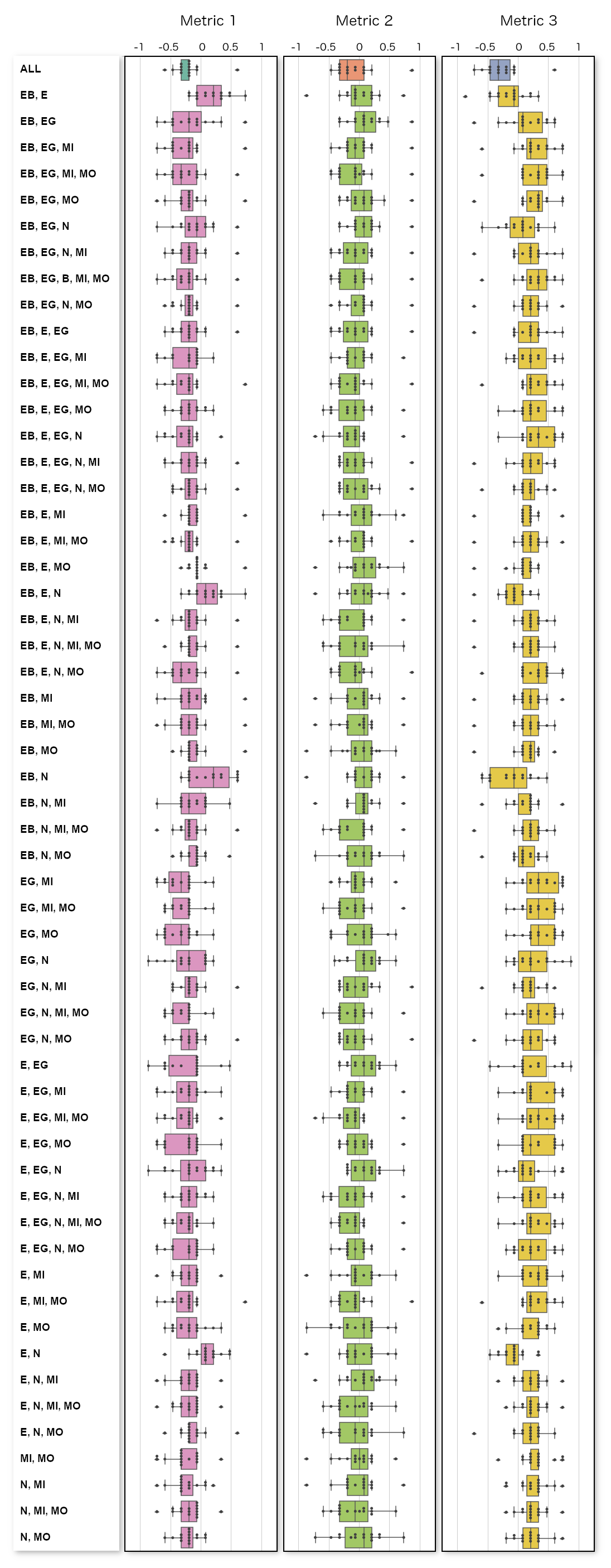}
\caption{\benjamin{Kendall's correlation coefficient between the AUC of FLL approaches and FER accuracy, in absence of head movements, w.r.t. the subsets of landmarks considered: EG - edges, EB - eyebrows, E - eyes, N - nose, MI - inner mouth, MI - outer mouth, ALL - all landmarks. The AUC is based on three metrics: Metric1 (traditional error metric), Metric2 (spatio-temporal stabilisation error based on the distance), Metric3 (spatio-temporal stabilisation error based on the angle).}}
\label{kendallsets}
\end{figure}

\begin{figure}[h!]
\centering
\includegraphics[width=\columnwidth]{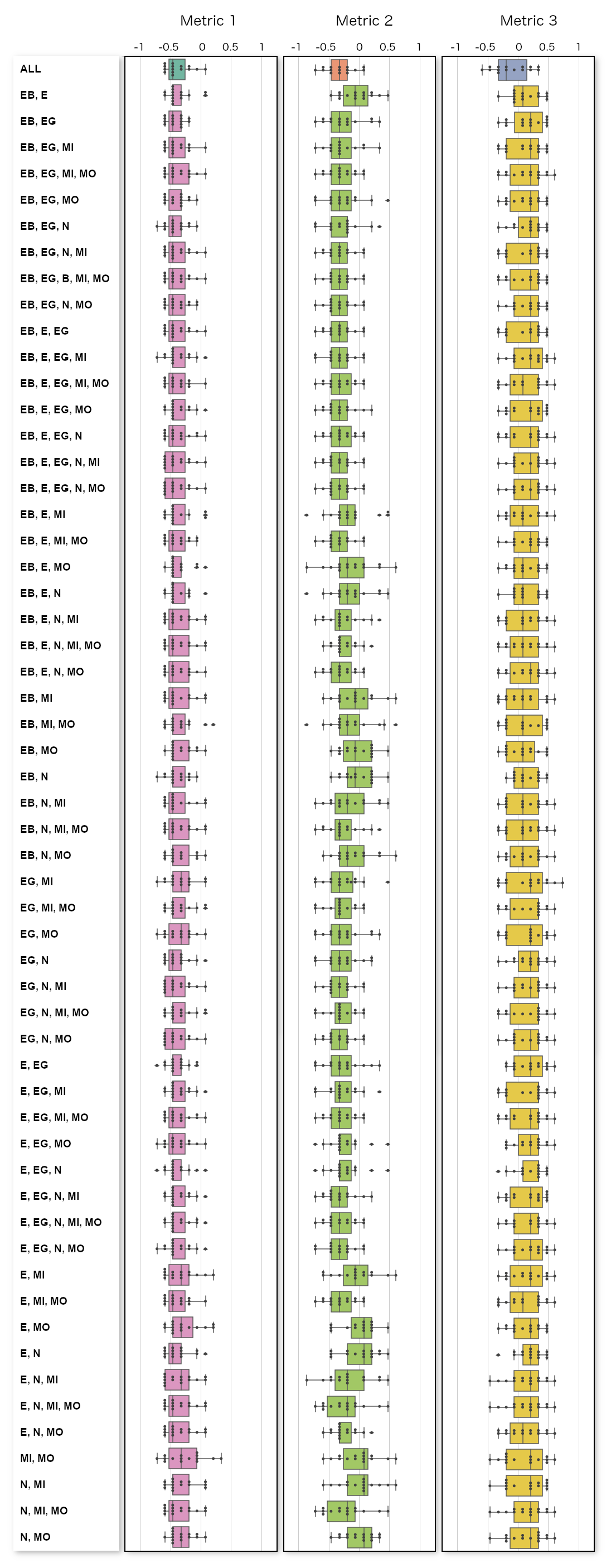}
\caption{\benjamin{Kendall's correlation coefficient between the AUC of FLL approaches and FER accuracy, in presence of head movements, w.r.t. the subsets of landmarks considered: EG - edges, EB - eyebrows, E - eyes, N - nose, MI - inner mouth, MI - outer mouth, ALL - all landmarks. The AUC is based on three metrics: Metric1 (traditional error metric), Metric2 (spatio-temporal stabilisation error based on the distance), Metric3 (spatio-temporal stabilisation error based on the angle).}}
\label{spearmansets}

\end{figure}

\benjamin{Figures~\ref{kendallsets} and \ref{spearmansets} show the correlation values measured when considering combinations of landmark subsets covering jointly several facial regions at a time}\pierre{, for each of the metrics mentioned in Section~\ref{metrics}. Correlations are measured in two settings: in the absence of head movements (based on FER accuracy values from Table~\ref{exp:tab3}) and in their presence (based on FER accuracy values from Table~\ref{exp:tab2}).}

\begin{figure}[h!]
\centering
\includegraphics[width=\columnwidth]{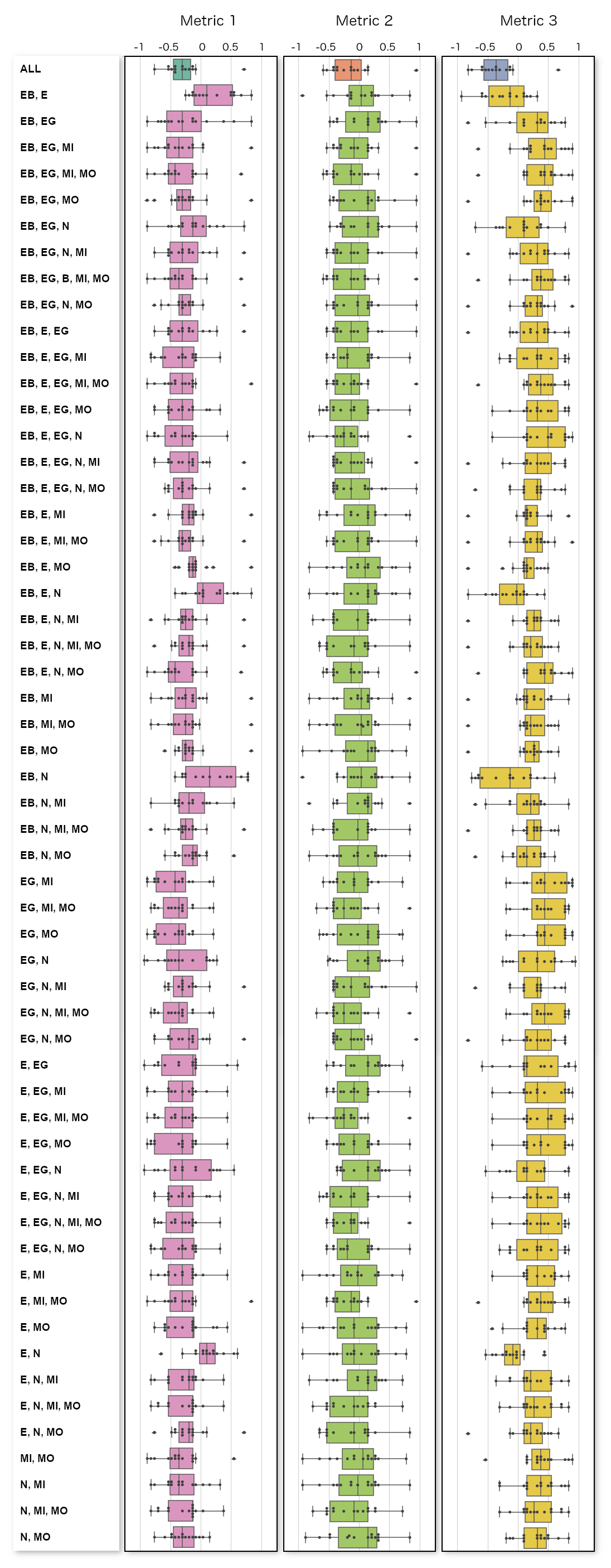}
\caption{\benjamin{Spearman's correlation coefficient between the AUC of FLL approaches and FER accuracy, in absence of head movements, w.r.t. the subsets of landmarks considered: EG - edges, EB - eyebrows, E - eyes, N - nose, MI - inner mouth, MI - outer mouth, ALL - all landmarks. The AUC is based on three metrics: Metric1 (traditional error metric), Metric2 (spatio-temporal stabilisation error based on distance), Metric3 (spatio-temporal stabilisation error based on angle).}}
\end{figure}

\begin{figure}[h!]
\centering
\includegraphics[width=\columnwidth]{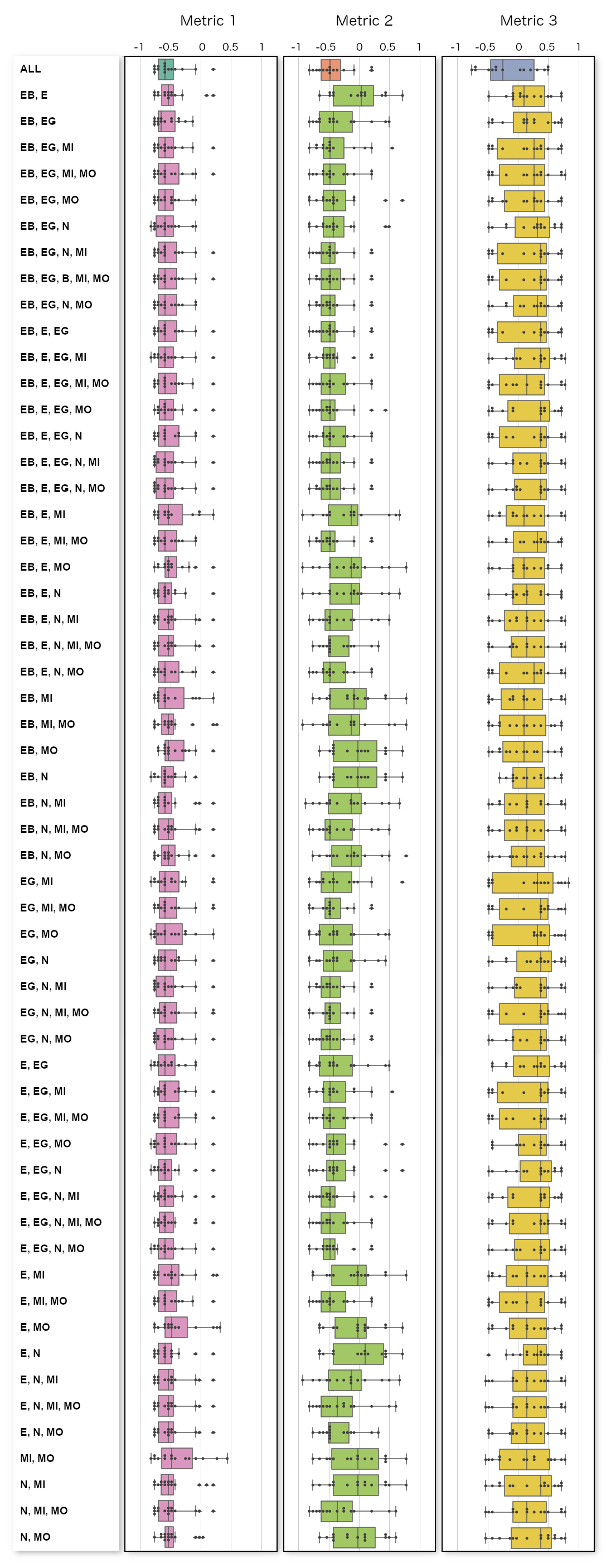}
\caption{\benjamin{Spearman correlation between the AUC of FLL approaches and FER accuracy, in presence of head movements, w.r.t. the subsets of landmarks considered: EG - edges, EB - eyebrows, E - eyes, N - nose, MI - inner mouth, MI - outer mouth, ALL - all landmarks. The AUC is based on three metrics: Metric1 (traditional error metric), Metric2 (spatio-temporal stabilisation error based on the distance), Metric3 (spatio-temporal stabilisation error based on the angle).}}
\end{figure}

\end{document}